\documentclass[10pt,journal,compsoc]{IEEEtran}

\ifCLASSOPTIONcompsoc
  \usepackage[nocompress]{cite}
\else
  \usepackage{cite}
\fi

\ifCLASSINFOpdf
\else
\fi

\usepackage{algorithmic}
\usepackage{array}
\usepackage{amsfonts}

\usepackage{epsfig}
\usepackage{graphicx}
\usepackage{amsmath}
\usepackage{amssymb}
\usepackage{bm}
\usepackage{cases}

\usepackage[pagebackref=true,breaklinks=true,letterpaper=true,colorlinks,bookmarks=false]{hyperref}

\usepackage[usenames, dvipsnames]{xcolor}

\usepackage{multirow}
\usepackage{algorithm}
\usepackage{algorithmic}
\usepackage{amsfonts}
\usepackage{mathtools,amssymb}
\usepackage{fixltx2e}
\usepackage{array}
\usepackage{setspace}
\usepackage{caption} 

\usepackage{bbold}

\usepackage{array}

\usepackage{amsmath}
\newcommand{\eqn}[1]{Equation~#1}
\newcommand{\fig}[1]{Figure~#1}
\newcommand{\tab}[1]{Table~#1}
\newcommand{\sect}[1]{Section~#1}

\newcommand{\etal}{\textit{et al. }}

\usepackage{expl3}

\ifCLASSOPTIONcompsoc
  \usepackage[caption=false,font=footnotesize,labelfont=sf,textfont=sf]{subfig}
\else
  \usepackage[caption=false,font=footnotesize]{subfig}
\fi
\usepackage{fixltx2e}
\usepackage{stfloats}

\ifCLASSOPTIONcaptionsoff
  \usepackage[nomarkers]{endfloat}
 \let\MYoriglatexcaption\caption
 \renewcommand{\caption}[2][\relax]{\MYoriglatexcaption[#2]{#2}}
\fi

\usepackage{url}

\begin{document}

\title{MonoGRNet: A General Framework for Monocular 3D Object Detection}

\author{Zengyi Qin,
        Jinglu Wang and 
        Yan Lu
\IEEEcompsocitemizethanks{
\IEEEcompsocthanksitem 
Z. Qin is with Massachusetts Institute of Technology, Cambridge, MA, 02139. This work was done when Z. Qin was an intern at Microsoft Research Asia. Email: qinzy@mit.edu

\IEEEcompsocthanksitem 
J. Wang and Y. Lu are with Microsoft Research Asia, Beijing, China, 100080. Email: \{jinglwa, yanlu\}@microsoft.com 
}
}

\IEEEtitleabstractindextext{
\begin{abstract}
Detecting and localizing objects in the real 3D space, which plays a crucial role in scene understanding, is particularly challenging given only a monocular image due to the geometric information loss during imagery projection. We propose MonoGRNet for the amodal 3D object detection from a monocular image via geometric reasoning in both the observed 2D projection and the unobserved depth dimension. MonoGRNet decomposes the monocular 3D object detection task into four sub-tasks including 2D object detection, instance-level depth estimation, projected 3D center estimation and local corner regression. The task decomposition significantly facilitates the monocular 3D object detection, allowing the target 3D bounding boxes to be efficiently predicted in a single forward pass, without using object proposals, post-processing or the computationally expensive pixel-level depth estimation utilized by previous methods. In addition, MonoGRNet flexibly adapts to both fully and weakly supervised learning, which improves the feasibility of our framework in diverse settings. Experiments are conducted on KITTI, Cityscapes and MS COCO datasets. Results demonstrate the promising performance of our framework in various scenarios.

\end{abstract}

}

\maketitle

\IEEEdisplaynontitleabstractindextext
\IEEEpeerreviewmaketitle

\IEEEraisesectionheading{\section{Introduction}\label{sec:introduction}}

\IEEEPARstart{A} crucial task in scene understanding is 3D object detection, which aims to predict the amodal 3D bounding boxes of objects from input sensory data such as LiDAR point clouds and images. Compared to LiDAR based~\cite{yang2018pixor,zhou2018voxelnet,lang2019pointpillars} and multi-view based~\cite{chen2017multiview, qi2017frustum, hu2018joint} 3D object detectors, monocular image-based methods~\cite{qin2019monogr, chabot2017deepmanta, mousavian20173dbox, You2019PseudoLiDARAD, wang2019pseudo, Weng2019pseudolidar, ding2019d4lcn, Jrgensen2019iouloss, ku2019monocular, chen2016monocular, he2019mono3d++} only take a single-view RGB image as input in inference, which means they have lower requirement on sensors and are less expensive in real-world implementation. If achieving satisfactory detection performance, they can become an important module in mobile robot perception. 
It would be more attractive if the monocular 3D detection can be learned without 3D labels, which require intensive labors.
However, monocular 3D object detection is an ill-posed problem due to the depth information loss in 2D image planes, let alone the challenges when 3D annotations are not offered.

To accurately detect and localize the objects in 3D using a monocular image, recent approaches~\cite{You2019PseudoLiDARAD, Weng2019pseudolidar, wang2019pseudo, Cai2020heightguided} have been proposed to first predict the pixel-level depth and convert the monocular image to 3D point cloud representations, then apply the well-developed LiDAR or multi-view based 3D object detectors to the point cloud. These methods with hybrid structures can achieve impressive detection accuracy, but also have inevitable limitations. 
They introduce additional expensive computational cost for predicting high-resolution depth maps from images, making them hardly feasible in mobile platforms where the energy and computing resource are limited. 
Furthermore, pixel-level depth estimation does not aim at predicting depth of objects of targeting classes but all pixels in the whole images. The uncertainty in irrelevant or unreliable pixels could bring precision loss into the final 3D object detection.
Apart from the pixel-level depth based approaches, there is another stream of methods~\cite{mousavian20173dbox, Liu2020SMOKESMKeypoint, Barabanau2019Monocular3OKeypoint, Jrgensen2019iouloss, Li2020RTM3DRM} that first predict the sparse 2D representations such as keypoints and 2D bounding boxes, then utilize optimization to fit a 3D bounding box, which can be very efficient. However, when the object is truncated by the image boundaries, the sparse 2D representations are partly missing, which impose significant challenges to fitting the 3D bounding box. In light of this, we will not rely on post-processing in our object detection framework. In addition, we hope that our framework can be free of object proposals to reduce the computational burden and improve the simplicity and generality.

\begin{figure*}[t]
	\centering 
	\includegraphics[width=1\linewidth]{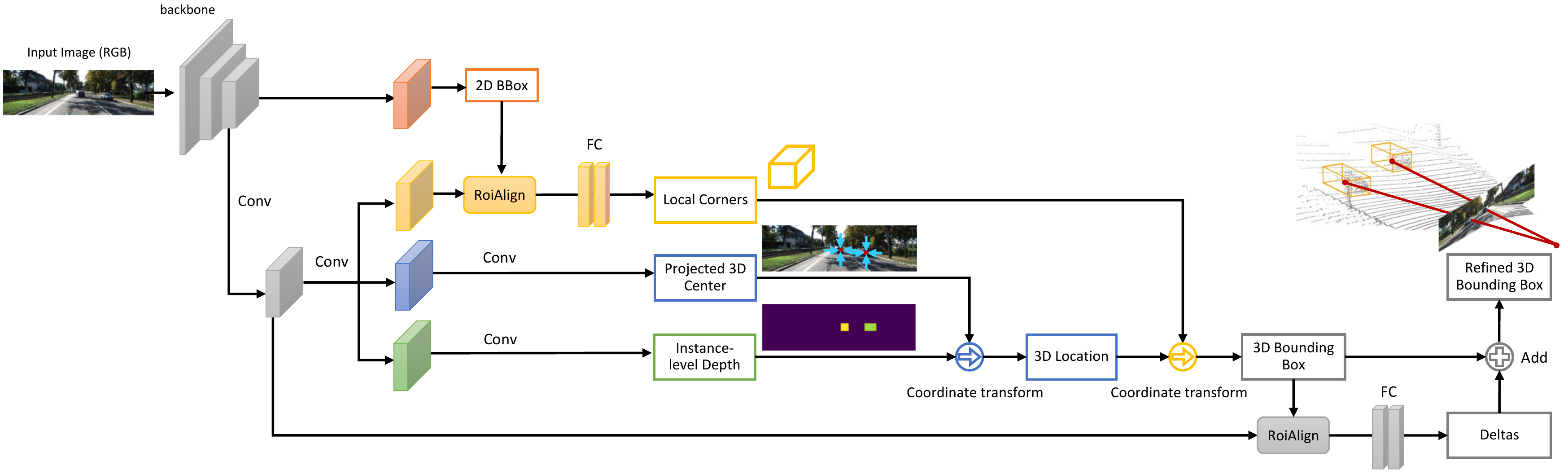}
	\caption{\textbf{Network structure.} 
	Our monocular 3D object detector consists of four sub-networks for 2D detection (\textcolor[RGB]{237,125,49}{brown}), 
	instance-level depth estimation (\textcolor[RGB]{112,173,71}{green}), projected 3D center 
	estimation (\textcolor[RGB]{91,155,213}{blue}) and local corner regression (\textcolor[RGB]{255,192,0}{yellow}). The output of the four sub-networks are combined to produce the 3D bounding boxes, which are refined to give the final outputs. Best viewed in color. It should be noticed that this overview figure only shows the inference stage. The training is achieved via either fully (see \sect{\ref{sect:full_train}}) or weakly (see \sect{\ref{sect:weak_train}}) supervised learning.}
	\label{fig:net}
\end{figure*}

In this paper, we present MonoGRNet, a general framework for learning \textbf{Mono}cular 3D object detection by \textbf{G}eometric \textbf{R}easoning. Taking a monocular RGB image as input, MonoGRNet predicts the 3D bounding boxes of objects in a single forward pass and can be implemented in a straightforward way. 
The proposed framework decouples the 3D object detection task into four progressive sub-tasks that are easy to solve using a monocular image. The four sub-tasks are 2D object detection, instance-level depth estimation, projected 3D center estimation and local corner regression, which are solved in parallel in the proposed unified network shown in \fig{\ref{fig:net}}. The network starts from 2D perception and then extends the geometric reasoning into 3D space. The instance-level depth estimation is crucial for bridging the 2D-to-3D gap, which is different from the computationally expensive pixel-level depth estimation utilized by many previous methods. Instance-level depth is defined as the depth of the 3D bounding box center, which can be interpreted as the depth of an object. Using the predicted instance-level depth, we backproject the estimated projected 3D center from the 2D image plane to 3D space to obtain the 3D center location of the object. At the same time, the local corner regression provides the coordinates of the eight corners of the 3D bounding box with respect to its 3D center. 

The proposed framework is general for variable learning schemes since sub-tasks are decoupled loosely.
It can be extended from fully supervised learning to a weakly supervised learning scenario, where the ground truth 3D bounding boxes are not available in training but we have the ground truth 2D bounding boxes instead. Labeling 2D bounding boxes saves much labor compared to labeling 3D bounding boxes based on the investigation~\cite{papadopoulos2017extreme}. This helps to improve the feasibility and efficiency of applying monocular 3D object detectors to various scenarios. In addition, widely used 3D shape datasets, such as PASCAL3D+~\cite{xiang_wacv14} and ShapeNet~\cite{Chang2015ShapeNetAI}, and their corresponding images make it easy to learn view angles of centered objects in a local extent. We present a geometric-guided method to learn the 3D location from labeled 2D bounding boxes and unlabeled frames, as well as an object-centric transfer learning method to learn the local corner regression with easily accessible 3D shape datasets. As we will see, the task decomposition that we propose is a crucial enabler of the flexible extension to weakly supervised learning of monocualr 3D object detection. The network structure and loss functions mostly remain unchanged in such an extension.

Our experiments are conducted in three public datasets, including KITTI~\cite{geiger2012kitti}, CiteScapes~\cite{Cordts2016Cityscapes} and MS COCO~\cite{Lin2014Microsoft}. Notice that CiteScapes~\cite{Cordts2016Cityscapes} and MS COCO~\cite{Lin2014Microsoft} do not provide ground truth 3D bounding boxes. Most of the existing 3D object detectors require full supervision, so these two datasets have not gained attention of the 3D object detection community. Our quantitative results on KITTI~\cite{geiger2012kitti} and qualitative results on all the three datasets demonstrate the promising performance of the proposed framework and its strong generalization capability to diverse training and testing scenarios. In summary, our contributions are:

\begin{itemize}
    \item We propose to decompose the monocular 3D object detection task into four sub-tasks including 2D object detection, instance-level depth estimation, projected 3D center estimation and local corner regression. Such formulation frees the detector from object proposals and computationally expensive pixel-level dense depth estimation used by previous methods.
    
    \item We propose a unified network to efficiently solve the four sub-tasks in parallel, which can be trained in an end-to-end fashion. We also present a method to train the network using ground truth 2D bounding boxes and easily accessible additional data when the ground truth 3D bounding boxes are unavailable.
    
    \item We conduct comprehensive experiments on the KITTI~\cite{geiger2012kitti}, Cityscapes~\cite{Cordts2016Cityscapes} and MS COCO~\cite{Lin2014Microsoft} datasets, demonstrating the advantage of the proposed framework in diverse scenarios. We also conduct ablation studies to examine the effectiveness of the crucial components in our framework.
\end{itemize}

A preliminary version of our MonoGRNet has been published~\cite{qin2019monogr} and gained attention. In this manuscript, we make the following improvement. 1) We simplify the structure of MonoGRNet to emphasize the most important concepts that we propose, and at the same time the quantitative performance is improved. 2) We extend the framework from fully supervised to weakly supervised learning, and present an effective method to train the network using ground truth 2D bounding boxes and easily accessible additional data when the ground truth 3D bounding boxes are not available during training. 3) We provide extensive quantitative and qualitative experimental results to show the performance of our framework in different settings and examine the effectiveness of the key modules in ablation studies.

\section{Related Work}

\subsection{2D Object Detection}
2D object detection deep networks are extensively studied. Region proposal based methods~\cite{girshick2015fast, ren2017faster} generate impressive results but perform slowly due to complex multi-stage pipelines. Another group of methods~\cite{redmon2016yolo, redmon2017yolo9000, liu2016ssd, fu2017dssd} focusing on fast training and inferencing apply a single stage detection. Multi-net~\cite{teichmann2016multinet} introduces an encoder-decoder architecture for real-time semantic reasoning. Its detection decoder combines the fast regression in Yolo~\cite{redmon2016yolo} with the size-adjusting RoiAlign of Mask-RCNN~\cite{he2017mrcn}, achieving a satisfied speed-accuracy ratio. All these methods predict 2D bounding boxes of objects while none 3D geometric features are considered.

\subsection{3D object detection}
Existing methods range from single-view RGB~\cite{chen2016monocular, xu2018multifusion, chabot2017deepmanta, kehl2017ssd6d}, 
multi-view RGB~\cite{chen2017multiview,chen20153dop}, to RGB-D~\cite{qi2017frustum,song2016deep,liu2015higher}. When geometric information of the depth dimension is provided, the 3D detection task is much easier. MV3D~\cite{chen2017multiview} generates 3D object proposals from bird's eye view maps given LiDAR point clouds, and then fuses features in RGB images, LiDAR front views and bird's eye views to predict 3D boxes. AVOD~\cite{hu2018joint} fuses the RGB and LiDAR information in the region proposal stage to reduce the missed detections. Given RGB-D data, F-PointNet~\cite{qi2017frustum} extrudes 2D region proposals to a 3D viewing frustum
 and then segments out the point cloud of interest object. Recently proposed state-of-the-art 3D object detectors include STD~\cite{yang2019std}, Part-A\^{}2 Net~\cite{shi2019part} and UberATG-MMF~\cite{liang2019multi}.

The most related approaches to ours are using a monocular RGB image. Information loss in the depth dimension significantly increases the task's difficulty. Performances of state-of-the-art such methods still have large margins to RGB-D and multi-view methods. Mono3D~\cite{cvpr16chen} leverages segmentation mask and contextual information to generate 3D object proposals. Mono3D++~\cite{he2019mono3d++} exploits pseudo 3D keypoints and shapes prior to localizing objects in 3D by minimizing matching loss. \cite{hu2018joint} combines monocular 3D detection with tracking in autonomous scenarios. MonoGRNet~\cite{qin2019monogr} proposes instance-level depth estimation to extend the 2D perception to 3D reasoning without reluctant intermediate representations. MonoPSR~\cite{ku2019monocular} leverages shape reconstruction from monocular images to facilitate 3D detection. Another line of research~\cite{wang2019pseudo, You2019PseudoLiDARAD} uses monocular images to generate pseudo point cloud, which is passed to existing point cloud based 3D detectors. Nevertheless, since extensive 3D labels are difficult to obtain in practice, the fully supervised methods have limitations in real-world applications. 

\subsection{Monocular Depth Estimation.}
Pixel-level depth estimation networks~\cite{wang2019pseudo, fu2018deep, ren2019deep} have been proposed. However, when regressing the pixel-level depth, the loss function takes into account every pixel in the depth map and treats them without significant difference. In a common practice, the loss values from each pixel are summed up as a whole to be optimized. Nevertheless, there is a likelihood that the pixels lying in an object are much fewer than those lying in the background, and thus the low average error does not indicate the depth values are accurate in pixels contained in an object. In addition, dense depths are 
often estimated from disparity maps that may produce large errors at far regions, which may downgrade the 3D localization performance drastically. Different from the pixel-level depth estimation methods, we propose an instance-level depth estimation (IDE) network that predicts the 3D center depth of objects. IDE does not require the densely labeled pixel-level depth for training and avoids the computationally expensive pixel-level monocular depth estimation in testing.

Instance-level depth has been studied in \cite{afifi2016object, zhang2015monocular, lee2019instance}. In \cite{afifi2016object}, the depths of objects are regressed via a single fully convolutional neural network. In \cite{zhang2015monocular}, the instance depth ordering is jointly learned with instance segmentation. In \cite{lee2019instance}, the authors proposed a self-supervised method to learn instance depth from video sequences. To the best of our knowledge, our MonoGRNet is the first to introduce instance-level depth estimation in the context of monocular 3D object detection. The instance-level depth enables the back-projection from 2D to 3D to obtain the 3D center locations of targeted objects.

\subsection{Weakly supervised object detection}
Most existing studies focus on 2D object detection, while weakly supervised 3D detection has not been extensively explored.
\cite{han2015object} starts by inferring the geometric structure implied in low-level and middle-level features to describe the objects of interest, then learns from high-level features by iterative training to detect objects in 2D. \cite{sangineto2019self} trains the detection network by iteratively selecting a subset of bounding boxes that are the most reliable. ACol~\cite{zhang2018adversarial} utilizes the output heatmaps for complementary learning. \cite{zhang2018zigzag} trains the weakly supervised localization network first on easy examples and then on hard ones. \cite{tang2018pcl} localizes objects by clustering. Different from these previous studies on 2D detection, we aim at bridging the gap between weakly supervised learning and 3D object detection.

\subsection{Transfer learning on object detection}
One popular usage of transfer learning in object detection is model parameter initialization, where the networks trained on large-scale image recognition datasets are used as CNN backbones in object detectors~\cite{li2019stereo, qin2019tlnet, braun2016pose, pathak2017learning, dai2016r}. Lim~\etal proposed to borrow examples from existing classes to learn to detect objects from unseen classes~\cite{lim2011transfer}. Lampert~\etal propose to use high-level attributes to detect object classes, making it easier to transfer to unseen classes~\cite{lampert2009learning}. Tang~\etal incorporate the visual and semantic similarities of objects in the transferring process. An important step in transfer learning is narrowing the gap between the source and the target dataset. Previous work has focused on domain adaptation~\cite{gustafsson2018automotive, hoffman2014lsda, spinello2012leveraging} at an image level, while we take a different route by ignoring the background and targeting at the object level, saving consideration on the gap between the source and target datasets in terms of the object sizes and background scenes in transfer learning.

\section{Problem Statement}
\label{sect:problem_statement}

\begin{figure}[t]
	\centering
	\includegraphics[width=1\linewidth]{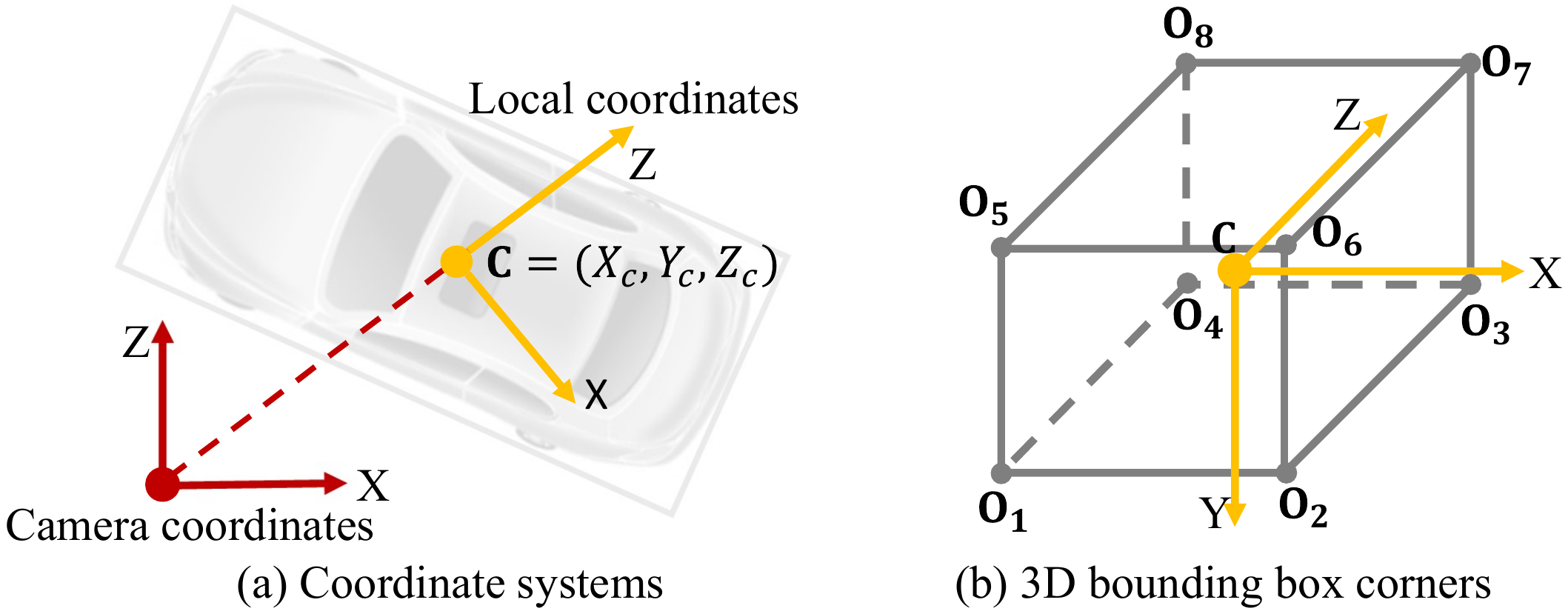}
	\caption{\textbf{Notations.} (a) illustrates the relationship between the camera coordinate system and the local coordinate system from the bird's eye view. (b) shows the 3D bounding box corners defined in the local coordinate system.}
	\label{fig:notation}
\end{figure}

Given a monocular RGB image, the objective is to detect and localize objects of specific classes in the 3D space. A target object is represented by a class label and a 3D bounding box, which bounds the complete object regardless of occlusion or truncation. A 3D bounding box $B_{3D}$ is defined by a 3D center $\mathbf{C}=(X_c,Y_c,Z_c)$ in global context and eight corners $\mathcal{O}=\{\mathbf{O}_k\}, k=1,...,8$, related to local context. \fig{\ref{fig:notation}} shows the notations. The 3D location $\mathbf{C}$ is defined in the camera coordinate frame and the local corners $\mathcal{O}$ are in a local coordinate frame whose origin is $\mathbf{C}$. It is clear that $\sum_{k=1}^8 \mathbf{O}_k = \mathbf{0}$
due to symmetry. 

\textbf{Fully supervised} learning of monocular 3D object detection means the fully annotated 3D bounding boxes $B_{3D}$ are provided for all objects of interest in the dataset. For \textbf{weakly supervised} learning, we assume the ground truth $B_{3D}$ is inaccessible throughout training, while the ground truth 2D bounding boxes $B_{2D}$ are available. The $B_{2D}$ provides much weaker supervision than $B_{3D}$, since the 3D information is almost lost. It should be noticed that previous work~\cite{han2015object, sangineto2019self, zhang2018adversarial, zhang2018zigzag, tang2018pcl} on weakly supervised object detectors focused on learning from image classification labels to predict the $B_{2D}$, while we assume that we already have the labeled $B_{2D}$ and focus on learning to predict $B_{3D}$. It is not impossible to learn from image-level classification labels rather than $B_{2D}$ to predict $B_{3D}$, but that is beyond the scope of this paper.

\section{Method}

We propose MonoGRNet, a general framework for monocular 3D object detection. MonoGRNet takes a monocular image as input and outputs the 3D bounding boxes of objects in a single forward pass, and is free of the object proposal stage and the computationally expensive pixel-level depth prediction. The framework adapts to both fully supervised and weakly supervised learning. Since directly predicting 3D bounding boxes from a 2D monocular image could be difficult due to the dimension reduction, we propose to decompose the task into four progressive sub-tasks that are easier to solve using a monocular image, even when the ground truth 3D bounding boxes are not available. The sub-tasks are (1) 2D object detection, (2) instance-level depth estimation, (3) projected 3D center estimation and (4) local corner regression. The 3D bounding box prediction result can be directly derived by combining the output of the four sub-tasks. In this section, we first describe the four sub-tasks and the network structure, then detail the learning process in fully and weakly supervised scenarios.

\subsection{Task Decomposition}

\begin{figure}
	\centering
	\includegraphics[width=1\linewidth]{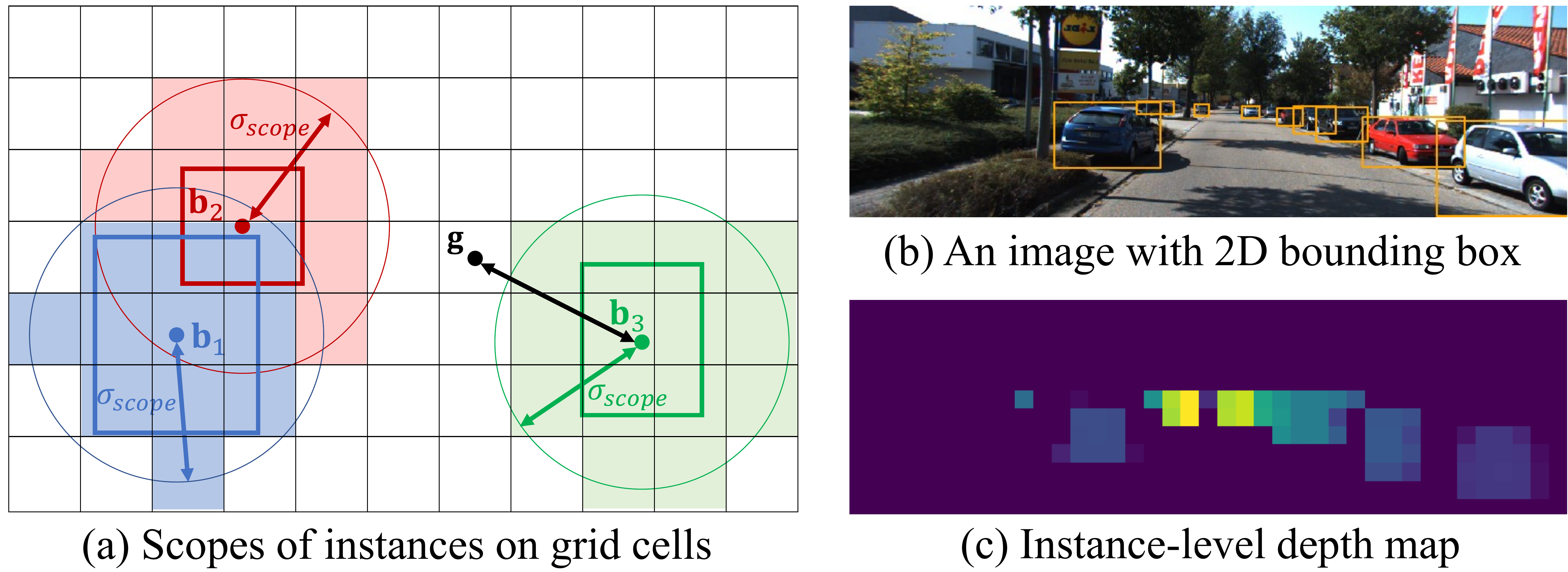}
	\caption{\textbf{Instance-level depth.} (a) Each grid cell $\mathbf{g}$ is assigned to a nearest object within a distance $\sigma_{scope}$ to the 2D bbox center $\mathbf{b}_i$. Objects closer to the camera are assigned to handle occlusion. Here $Z_c^1 < Z_c^2$. (b) An image with detected 2D bounding boxes. (c) Predicted instance depth for each cell.}
	\label{fig:instance_depth}
\end{figure}

The 3D bounding box $B_{3D}$ is the final goal of prediction. Before predicting the $B_{3D}$ of an object on the image, the network should be aware of the presence of the object. Therefore, the first task is \textbf{2D object detection}, which aims to predict the 2D bounding box $B_{2D}$ of the object and its class. $B_{2D}=(w_{2D}, h_{2D}, u_b, v_b)$ where $(w_{2D}, h_{2D})$ indicates the size of $B_{2D}$ and $(u_b, v_b)$ represents the center of $B_{2D}$ on the image plane. 

As is stated in \sect{\ref{sect:problem_statement}}, $B_{3D}$ is parameterized by a 3D center $\mathbf{C}=(X_c,Y_c,Z_c)$ in camera coordinates and eight corners $\mathcal{O}=\{\mathbf{O}_k\}, k=1,...,8$, defined in local coordinates. It can be difficult to directly regress $\mathbf{C}$ of an object from the image features because the image itself does not have explicit depth information. Recent works~\cite{You2019PseudoLiDARAD, wang2019pseudo} propose to first predict the depth of each pixel of the monocular image in order to convert the image to point cloud representations, which are fed to a point cloud based 3D object detector. The pixel-level depth prediction demands considerable computation resource, limiting its feasibility in diverse application scenarios such as mobile robot platforms. Moreover, the final prediction of 3D object detection is instance-level rather than pixel-level. The uncertainty in irrelevant background pixels could hamper the prediction accuracy. In light of this, we propose the \textbf{instance-level depth estimation} as the second task, which aims to predict the depth $Z_c$ of the 3D center $\mathbf{C}$ of the 3D bounding box. Examples of the predicted instance-level depths are illustrated in \fig{\ref{fig:instance_depth}}~(c). The third task is \textbf{projected 3D center estimation}, which will bring us closer to obtaining $\mathbf{C}=(X_c,Y_c,Z_c)$. The projected 3D center is a 2D point $\mathbf{c}=(u_c, v_c)$ defined as the projection of $\mathbf{C}$ on image. $X_c$ and $Y_c$ can be derived by:
\begin{align}
\label{eq:2d_3d}
X_c=(u_c-p_u) * Z_c / f_u, & \quad &
Y_c=(v_c-p_v) * Z_c / f_v
\end{align}
where $f_u$ and $f_v$ are the focal length along X and Y axes, $p_u$ and $p_v$ are coordinates of the principle point. These are a part of the intrinsic parameters of the camera and can be easily obtained when the camera is calibrated. \eqn{\ref{eq:2d_3d}} can be interpreted as back-projecting $\mathbf{c}$ from 2D to 3D using $Z_c$ and the camera parameters. We have demonstrated how to predict $\mathbf{C}$ by estimating $Z_c$ and $\mathbf{c}$. $\mathbf{c}$ is a 2D point on the image plane, so regressing $\mathbf{c}$ is much easier than regressing $X_c$ and $Y_c$.

The fourth task is local corner regression, which aims to estimate the eight corners $\mathcal{O}=\{\mathbf{O}_k\}, k=1,...,8$ of the 3D bounding box $B_{3D}$ in a local coordinate system described in \sect{\ref{sect:problem_statement}} and \fig{\ref{fig:notation}}. To summarize, the first task gives $B_{2D}$ and the object class, the second and third tasks give $\mathbf{C}$, and the forth task gives $\mathcal{O}$. $\mathbf{C}$ and $\mathcal{O}$ completely parameterize the $B_{3D}$, which is the final output. Due to the well-decomposed task, our framework can also be extended from fully supervised learning to weakly supervised learning, which will be detailed in \sect{\ref{sect:weak_train}}.

\subsection{Network Structure}
We design a unified end-to-end network to efficiently solve the four sub-tasks in parallel, which is illustrated in \fig{\ref{fig:net}}. The network takes a monocular image as input and outputs a set of $B_{3D}$ corresponding to the objects on the image. The four sub-tasks share the same backbone and only differ in head layers, which enable feature reuse and improve the inference efficiency. Compared to previous monocular 3D object detectors~\cite{You2019PseudoLiDARAD, wang2019pseudo, ku2019monocular, chen2016monocular}, the proposed network does not require the dense pixel-level depth prediction, instance segmentation or ground plane estimation, and is free of extensive object proposals~\cite{ren2017faster}.

The input image is divided into an $S_u \times S_v$ grid $\mathcal{G}$, where a cell is indicated by $\mathbf{g}$. The image is passed to a fully convolutional backbone network, whose output feature map is also of size $S_u \times S_v$, followed by the head layers of the sub-tasks. The head layers do not down-sample the feature maps, so the resolution remains $S_u \times S_v$. Each pixel in the feature map corresponds to a cell in the image grid, and will predict the nearest object on image. A single pixel in the head layer feature maps, namely a grid cell, can have multiple channels to regress multiple values.

In the 2D object detection branch, each grid cell $\mathbf{g}$ outputs the object classification probability $P^{\mathbf{g}}$ and the 2D bounding box $B_{2D}^{\mathbf{g}}=(w_{2D}^{\mathbf{g}}, h_{2D}^{\mathbf{g}}, u_b^{\mathbf{g}}, v_b^{\mathbf{g}})$, indicated by the superscript $\mathbf{g}$. We use softmax activation for $P^{\mathbf{g}}$ and no activation for $B_{2D}^{\mathbf{g}}$ in the last layer. The regression target for $(w_{2D}^{\mathbf{g}}, h_{2D}^{\mathbf{g}})$ is itself, while the regression target for $(u_b^{\mathbf{g}}, v_b^{\mathbf{g}})$ is $(\Delta_{u_b}^{\mathbf{g}}, \Delta_{v_b}^{\mathbf{g}})=(u_b^{\mathbf{g}}-u_g, v_b^{\mathbf{g}}-v_g)$, which is the residuals between the the central location of $B_{2D}^{\mathbf{g}}$ and $\mathbf{g}=(u_g, v_g)$.

In the instance-level depth estimation branch, each grid cell $\mathbf{g}$ regresses a $Z_c^{\mathbf{g}}$. In the projected 3D center estimation branch, each grid cell predicts a $\mathbf{c}^{\mathbf{g}}=(u_c^{\mathbf{g}}, v_c^{\mathbf{g}})$. The regression target is the residuals $(\Delta_{u_c}^{\mathbf{g}}, \Delta_{v_c}^{\mathbf{g}})=(u_c^{\mathbf{g}}-u_g, v_c^{\mathbf{g}}-v_g)$. In the local corner regression branch, for each grid cell, we use the RoIAlign~\cite{he2017mrcn} to crop the features bounded by $B_{2D}^{\mathbf{g}}$ from the feature maps produced by the backbone network. Then the features are passed to fully connected layers to regresses eight 3D corners $\mathcal{O}^{\mathbf{g}}=\{\mathbf{O}_k^{\mathbf{g}}\}, k=1,...,8$. The last layers for the three tasks are without an activation function.

The 3D center $\mathbf{C}^{\mathbf{g}}$ is easily calculated from \eqn{\ref{eq:2d_3d}} using $Z_c^{\mathbf{g}}$ and $\mathbf{c}^{\mathbf{g}}$. $\mathbf{C}^{\mathbf{g}}$ and $ \mathbf{O}_k^\mathbf{g}$ defines a 3D bounding box predicted at grid cell $\mathbf{g}$. Finally, we project the 3D box onto the image plane, obtain a 2D box and use RoIAlign~\cite{he2017mrcn} to extract features bounded by the 2D box from the feature maps of the backbone network. Then we pass the features to fully connected layers to regress $\Delta{\mathbf{C}}^\mathbf{g}$ and $\Delta{\mathbf{O}_k^\mathbf{g}}$ to refine the prediction. $\mathbf{C}^\mathbf{g}+\Delta{\mathbf{C}}^\mathbf{g}$ and $\mathbf{O}_k^\mathbf{g}+\Delta{\mathbf{O}_k^\mathbf{g}}, k=1,...,8$ represent the refined 3D bounding box, which is the final output of grid cell $\mathbf{g}$. Note that the 3D bounding box refinement is a complementary stage and is not contained in the four fundamental sub-tasks. For a single image, the network gives $S_u \times S_v$ 3D bounding boxes $B_{3D}$ in total. The final prediction of the network is obtained by applying non-maximum suppression to the $B_{3D}$.

\subsection{Fully Supervised Learning}
\label{sect:full_train}
In fully supervised learning, the ground truth $B_{3D}$ is provided, and the corresponding $B_{2D}$ is obtained by projecting the $B_{3D}$ to the 2D image plane. Here we formally formulate the loss functions of the sub-task under fully supervision. In this subsection, to distinguish between the network prediction and ground truth, the ground truth is modified by the $\hat{(\cdot)}$ symbol in the loss functions.

We start from assigning ground truth to each cell $\mathbf{g}$ in the $S_u \times S_v$ grid. A ground truth object is assigned to a cell $\mathbf{g}$ if the distance between the 2D bounding box and $\mathbf{g}$ is less than $\sigma_{scope}$. If a $\mathbf{g}$ is assigned to multiple objects, we only choose the object with the smallest instance-level depth $Z_c$. In a frame, some $\mathbf{g}$ does not have any ground truth because they are too far from the objects. These $\mathbf{g}$ are considered as background, while the remaining is foreground. We use $\mathbf{FG}$ to denote the set of foreground $\mathbf{g}$.

The classification output is trained using softmax cross entropy ($CE$) loss and the 2D bounding box regression is trained by L1 distance loss: 
\begin{align}
\mathcal{L}_{P} = &\sum_{\mathbf{g} \in \mathcal{G}} CE(P^{\mathbf{g}}, \hat{P}^{\mathbf{g}}) \nonumber \\
\mathcal{L}_{B_{2D}} = &\sum_{\mathbf{g} \in \mathbf{FG}} (|w_{2D}^{\mathbf{g}} - \hat{w}_{2D}^{\mathbf{g}}| + |h_{2D}^{\mathbf{g}} - \hat{h}_{2D}^{\mathbf{g}}| \nonumber \\
& \quad \quad + |\Delta_{u_b}^{\mathbf{g}} - \hat{\Delta}_{u_b}^{\mathbf{g}}| + |\Delta_{v_b}^{\mathbf{g}} - \hat{\Delta}_{v_b}^{\mathbf{g}}|)
\end{align}
The instance-depth estimation, projected 3D center estimation and local corner regression are trained with L1 loss:
\begin{align}
&\mathcal{L}_{Z_c} = \sum_{\mathbf{g} \in \mathbf{FG}} |Z_{c}^{\mathbf{g}} - \hat{Z}_{c}^{\mathbf{g}}| \label{eqn:loss_z}\\
&\mathcal{L}_{\mathbf{c}} = \sum_{\mathbf{g} \in \mathbf{FG}} (|\Delta_{u_c}^{\mathbf{g}} - \hat{\Delta}_{u_c}^{\mathbf{g}}| + |\Delta_{v_c}^{\mathbf{g}} - \hat{\Delta}_{v_c}^{\mathbf{g}}|) \label{eqn:loss_c} \\
&\mathcal{L}_{\mathcal{O}} = \sum_{\mathbf{g} \in \mathbf{FG}} \sum_{k=1}^8 |\mathbf{O}_{k}^{\mathbf{g}} - \mathbf{\hat{O}}_{k}^{\mathbf{g}}| \label{eqn:loss_o}
\end{align}
The 3D bounding box refinement is trained with L1 loss:
\begin{align}
&\mathcal{L}_{\Delta \mathbf{C}} = \sum_{\mathbf{g} \in \mathbf{FG}} |\Delta \mathbf{C}^{\mathbf{g}} - (\hat{\mathbf{C}}^{\mathbf{g}} - \mathbf{C}^{\mathbf{g}})| \label{eqn:loss_delta_c}\\
& \mathcal{L}_{\Delta \mathcal{O}} = \sum_{\mathbf{g} \in \mathbf{FG}} \sum_{k=1}^8 |\Delta \mathbf{O}_{k}^{\mathbf{g}}- (\mathbf{\hat{O}}_{k}^{\mathbf{g}} - \mathbf{O}_{k}^{\mathbf{g}})| \label{eqn:loss_delta_o}
\end{align}
Finally, we sum up the losses to produce the final loss function to be minimized.

\subsection{Weakly Supervised Learning}
\label{sect:weak_train}
In weakly supervised learning, we consider the scenario where the training set provides 2D bounding boxes instead of 3D bounding boxes, as is stated in \sect{\ref{sect:problem_statement}}. We also assume that 3D data such as point clouds or depth maps are not available in neither training nor testing. As the ground truth 2D bounding boxes are provided, we use the same loss functions to train the 2D object detection branch as is in \sect{\ref{sect:full_train}}. The main challenge is to learn the remaining three tasks. For instance-level depth estimation and projected 3D center estimation, we propose the geometry-guided learning method, which makes full use of the projective geometry to guide the learning process using 2D bounding boxes and unlabeled frames. For local corner regression, we propose the object-centric transfer learning method that can effectively transfer the knowledge from another easily accessible dataset to the target object detector to facilitate the learning process.

\subsubsection{Geometry-Guided Learning of 3D Location}
\label{sect:learning_weak_loc}

\begin{figure}
    \centering
    \includegraphics[width=\linewidth]{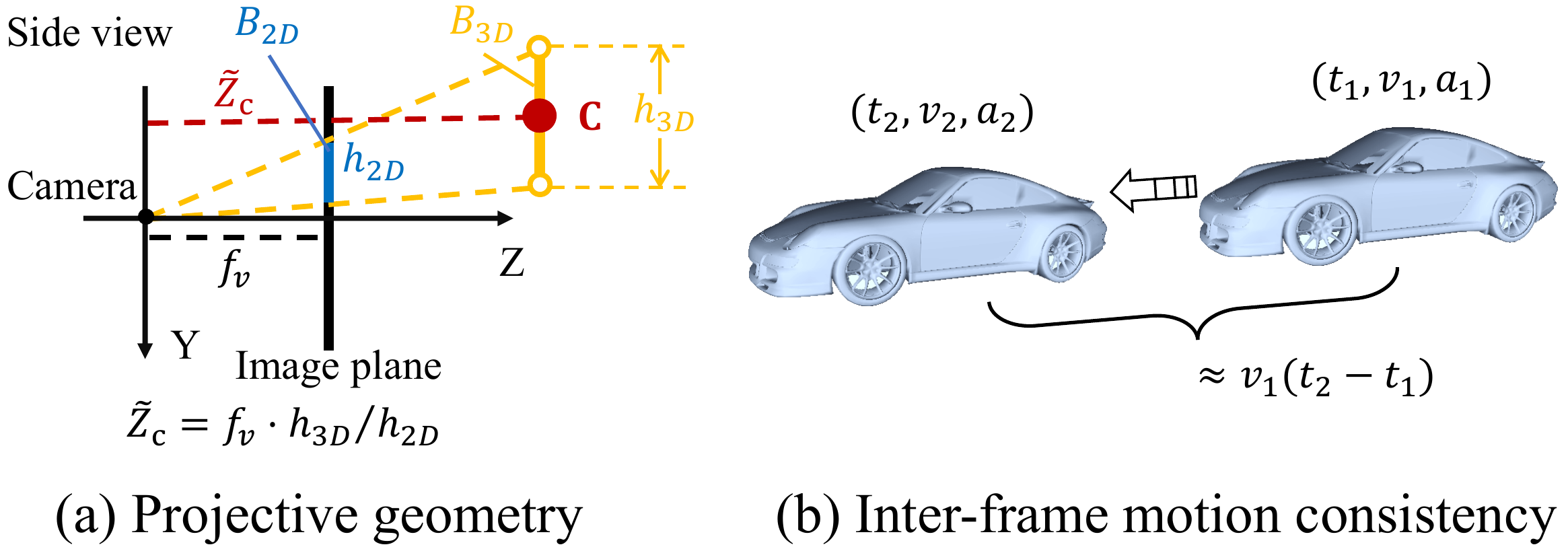}
    \caption{\textbf{Geometry-guided learning of 3D location.} (a) illustrates the projective geometry based on which we calculate $\widetilde{Z}_{c}$ as a pseudo ground truth of $Z_c$ to supervise the instance-level depth estimation. (b) shows the motion of the same object across neighbouring frames, where its acceleration should be no greater than a certain threshold. This acceleration constraint is imposed onto the network predictions to increase the 3D object localization performance.}
    \label{fig:geo_learn}
\end{figure}

We consider the second and third tasks, instance-level depth estimation and projected 3D center estimation. Their goal is to obtain the 3D location $\mathbf{C}$. Using ground truth 2D bounding boxes as supervision, we propose to leverage the projective geometry and unlabeled frames to learn the two tasks. Denote the prior height of the object as $h_{3D}$ and the camera focal length along the $v$-axis as $f_v$. Since we have the height $h_{2D}$ of the ground truth $B_{2D}$, we can calculate a rough instance-level depth $\widetilde{Z}_{c} = f_v \cdot h_{3D}~/~h_{2D}$ that is illustrated in \fig{\ref{fig:geo_learn}}~(a). We regard $\widetilde{Z}_{c}$ as a pseudo ground truth $Z_c$ and use it to replace $\hat{Z}_{c}$ in \eqn{\ref{eqn:loss_z}} to train the network. We remove the subscript $\mathbf{g}$ for simplicity. Also, we regard $\hat{\mathbf{c}}=(u_b, v_b)$, the center of the ground truth $B_{2D}$, as a pseudo ground truth to calculate the regression targets $\Delta_{u_c}^{\mathbf{g}}$ and $\Delta_{v_c}^{\mathbf{g}}$ in \eqn{\ref{eqn:loss_c}} to train the network. Given that the pseudo ground truth is only a rough approximation, we will refine the network predictions soon.

A rough 3D location $\mathbf{C}$ can be calculated from the estimated ${Z}_c$ and $\mathbf{c}$ of the grid cell using \eqn{\ref{eq:2d_3d}}. Then we refine this prediction by regressing $\Delta\mathbf{C} = (\Delta {X}_c, \Delta Y_c, \Delta {Z}_c)$, so that $\Delta{\mathbf{C}} + {\mathbf{C}}$ is closer to the real 3D location. Since the 3D ground truth is absent, we propose to utilize the first-order approximation of $\Delta\mathbf{C}$ to train the network. Here we assume we already have the local corners $\mathcal{O}$ and will explain how to obtain $\mathcal{O}$ in \sect{\ref{sect:object_centric}}. A coarse 3D bounding box ${B}_{3D}$ can be determined by ${\mathbf{C}}$ and $\mathcal{O}$. By projecting ${B}_{3D}$ to the image, we obtain a projected 2D bounding box, then we subtract it from the ground truth 2D bounding box to get $\Delta B_{2D} = (\Delta w_{2D}, \Delta h_{2D}, \Delta u_{v}, \Delta  v_{b})$. The approximated $\Delta\widetilde{\mathbf{C}}$ is formulated as:
\begin{align}
    \label{eq:apxdelta}
    &\Delta\widetilde{\mathbf{C}} = (\frac{\partial X_c}{\partial b_u} \Delta b_{u}, \frac{\partial Y_c}{\partial b_v} \Delta  b_{v}, \frac{\partial Z_c}{\partial h_{2D}} \Delta h_{2D})
\end{align}
where the partial derivatives are: 
\begin{align}
    \label{eq:devxyz}
    & \frac{\partial X_c}{\partial b_u} \approx \frac{\partial X_c}{\partial c_u} = \frac{Z_c}{f_u}, \quad
     \frac{\partial Y_c}{\partial b_v} \approx \frac{\partial Y_c}{\partial c_v} = \frac{Z_c}{f_v} \nonumber\\
    & \frac{\partial Z_c}{\partial h_{2D}} \approx \frac{\partial \widetilde{Z}_c}{\partial h_{2D}} = -\frac{f_v \cdot h_{3D}}{h_{2D}^{2}}
\end{align}
In \eqn{\ref{eqn:loss_delta_c}} we replace $(\hat{\mathbf{C}}^{\mathbf{g}} - \mathbf{C}^{\mathbf{g}})$ with $\Delta\widetilde{\mathbf{C}}$ to train the network. The previous method Deep3DBox~\cite{mousavian20173dbox} minimizes the re-projection discrepancy to obtain 3D locations, but it regards the optimization as post-processing and has to predict the 2D bounding boxes before calculating the 3D bounding boxes in inference. Our approach uses the 2D bounding boxes to endow the network with the ability of 3D location estimation directly from image inputs, which means the pipeline can predict the 3D bounding boxes in an end-to-end fashion.

If neighbouring image frames, such as video data, are available, we can impose another regularization based on a real-world kinematics prior that the acceleration of objects should be limited to a certain threshold. \fig{\ref{fig:geo_learn}}~(b) shows the motion of an object across frames. We formulate this acceleration constraint as:
\begin{align}
    & \mathcal{L}_{a} = \sum_{k=1}^{K} \sum_{n\in \mathbf{FN}_{k}} {\rm clip}(|{\bm{a}}_{n}^{k}| - \alpha_{a}, 0, \beta_{a}) \nonumber\\
    & {\bm{a}}_{n}^{k} = \frac{{\bm{v}}_{n}^{k} - {\bm{v}}_{n+1}^{k}}{t_{n} - t_{n+1}}, \quad
     {\bm{v}}_{n}^{k} = \frac{E_{\mathbf{FG}_{k, n}}[{\mathbf{C}}] - E_{\mathbf{FG}_{k, n+1}}[{\mathbf{C}}]}{t_{n} - t_{n+1}}
\end{align} 
where $\mathbf{FN}_{k}$ is the set of indices of the frames in which object $k$ is present, $\mathbf{FG}_{k, n}$ refers to the set of foreground cells of object $k$ in the $n^{th}$ frame, $\bm{a}_{n}^{k}$ and $\bm{v}_{n}^{k}$ are the instantaneous acceleration and velocity of the object relative to the camera, $t_{n}$ indicates the corresponding time. The gradient from $\mathcal{L}_{a}$ back-propagates to $Z_c$ and $\mathbf{c}$ through $\mathbf{C}$. $\mathcal{L}_{a}$ equals to zero if $|\bm{a}_{n}^{k}|$ is less than the threshold $\alpha_{a}$. $\beta_{a}$ is to clip the loss to avoid instability. We choose $\alpha_{a}=0.3$ and $\beta_{a}=3.0$ by grid search. We do not need the ground truth 2D bounding boxes in frame $2, 3, \cdots, N$, since the foreground region in these frames are estimated using the bounding boxes in the first frame and the inter-frame optical flow that is obtained using the off-the-shelf PWC-Net ~\cite{sun2018pwc}. Note that motion consistency is employed in training, while a single image is enough in inference. 

Note that the geometric-guided learning of 3D location detailed above requires the prior size (or at least the prior height) of the object and the camera intrinsics, including the principle point $(p_u, p_v)$ and focal lengths $(f_u, f_v)$. Each type of objects is assigned a prior size that equals to the average size of that type of objects. In terms of the camera parameters, since $(p_u, p_v)$ is always very close to the image center, when the camera intrinsics are unknown such as in MS COCO~\cite{Lin2014Microsoft} dataset, the principle point is set to be the image center. For $(f_u, f_v)$, we set $f_u$ to be 0.8 times the image width and $f_v = f_u$, which is also adopted by~\cite{mehta2018single}. If the users can provide an accurate size (height denoted as $\hat{h}_{3D}$) of the object and the real focal length $\hat{f}_v$ of the camera, then they can multiply the predicted instance-level depth by constant $\hat{h}_{3D} f_v / h_{3D} \hat{f}_v$ to obtain a more accurate prediction.

\subsubsection{Object-Centric Transfer Learning}
\label{sect:object_centric}

\begin{figure}
    \centering
    \includegraphics[width=\linewidth]{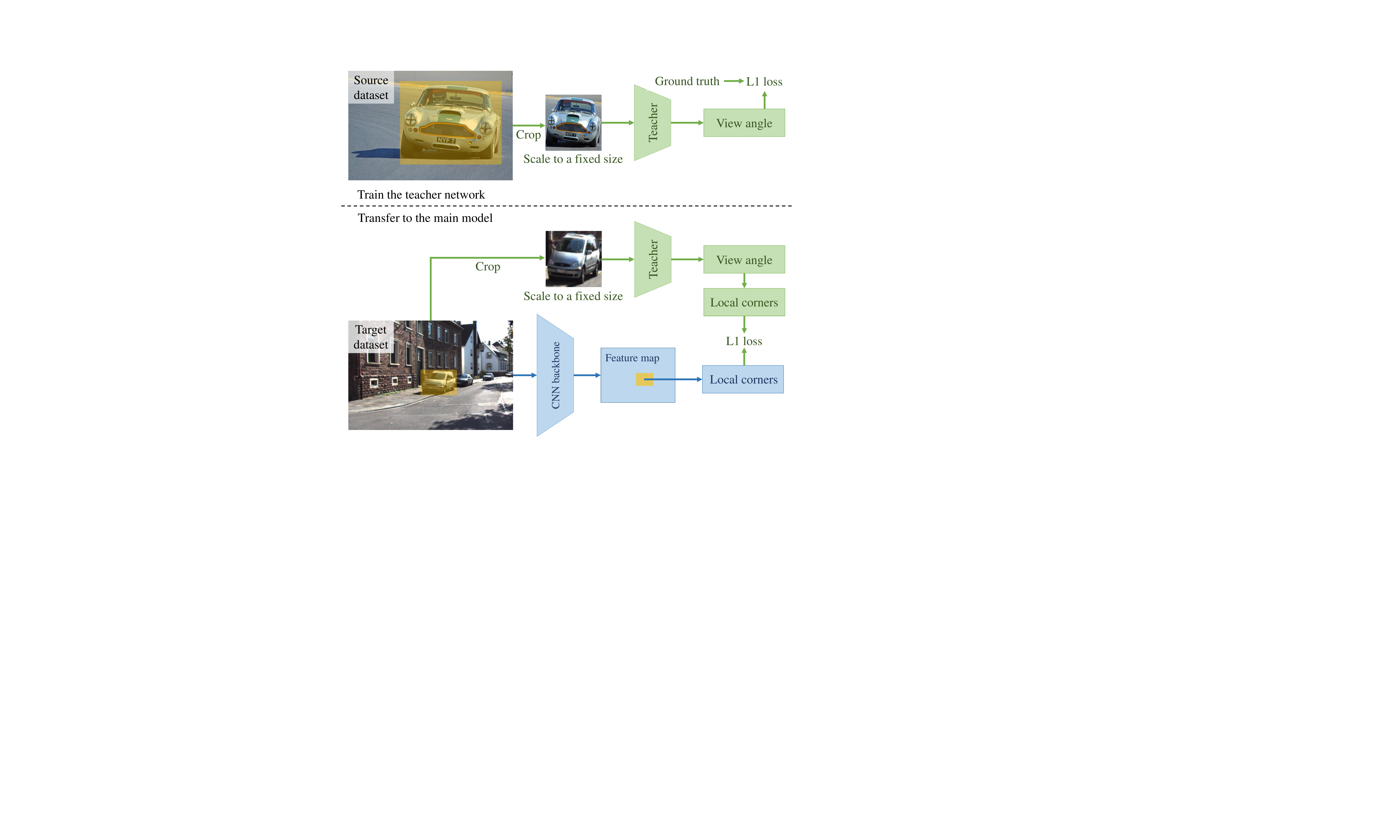}
    \caption{\textbf{Object-centric transfer learning for local corner estimation.} The blue part and green part are the abstraction of the student and teacher networks respectively. The student network is our monocular 3D object detector. This figure shows how the local corner regression branch is learned using easily accessible additional data that is denoted by the source dataset. The teacher network serves as a media transferring knowledge from the source to the target dataset, saving the arduous annotation on the target dataset. The teacher network only deals with local regions with objects of interest, rather than taking the whole image as input.}
    \label{fig:trans_learn}
\end{figure}
We consider the fourth task, local corner regression. The ground truth corners $\mathbf{O}_k$ are not contained in the labeled $B_{2D}$ and should be learned by introducing another source of knowledge. Note that the corners can be computed from the size and orientation of the $B_{3D}$. For simplicity, we use the prior size of each class of objects, so the unknown parameters are reduced to the orientation. It is noticed that there are many easily accessible datasets (e.g., PASCAL3D+~\cite{xiang_wacv14} and ShapeNet~\cite{Chang2015ShapeNetAI}) annotated with object view angles. We present an object-centric transfer learning method (see~\fig{\ref{fig:trans_learn}}) to use the additional data to train the local corner regression branch, which saves labor-intensive annotation on new datasets. We use PASCAL3D+~\cite{xiang_wacv14} as an example source dataset. It should be noted that only the ground truth view angle is required, which means the source dataset needs not to be annotated with the complete 6-DoF poses as is in PASCAL3D+~\cite{xiang_wacv14}.

The first step is training a teacher network on the source dataset~\cite{xiang_wacv14} using its ground truth view angle. The teacher regresses the cosine and sine values of the view angle. Full annotations of 6-DoF poses are not required. The most critical operation is cropping out the objects and scaling them to a fixed size (we use $\text{64}\times\text{64}$) before feeding to the network. In this way, we eliminate the interference of background distributions and object sizes, making each instance scale-invariant. The 2D bounding boxes are obtained via off-the-shelf object detectors~\cite{ren2017faster} if they are not labeled.

The second step is transferring the knowledge to our monocular 3D object detector. Using the ground truth 2D bounding box, we crop out the objects and scale them to the same size that the teacher is trained. The teacher predicts the view angle of each object online. Using the annotated 2D bounding box $B_{2D}=(w_{2D}, h_{2D}, u_b, u_p)$ and camera intrinsics, we convert the view angle to orientation. In our setting, only the orientation on the ground plane is considered. Let $\varphi$ be the view angle around the axis perpendicular to the ground plane, the orientation $\theta$ can be analytically calculated as $\theta = \varphi - arctan(\frac{u_b-u_p}{f_u})$, where $u_p$ is the horizontal principal point and $f_u$ is the horizontal focal length of the calibrated monocular camera. Then we use the orientation and the prior size to compute the approximated ground truth $\widetilde{\mathbf{O}}_k$ to supervise the targeted 3D detector. We replace $\hat{\mathbf{O}}_k$ with $\widetilde{\mathbf{O}}_k$ in \eqn{\ref{eqn:loss_o}} to train the network.

\section{Experiment}

\begin{figure}[t]
    \centering
    \includegraphics[width=\linewidth]{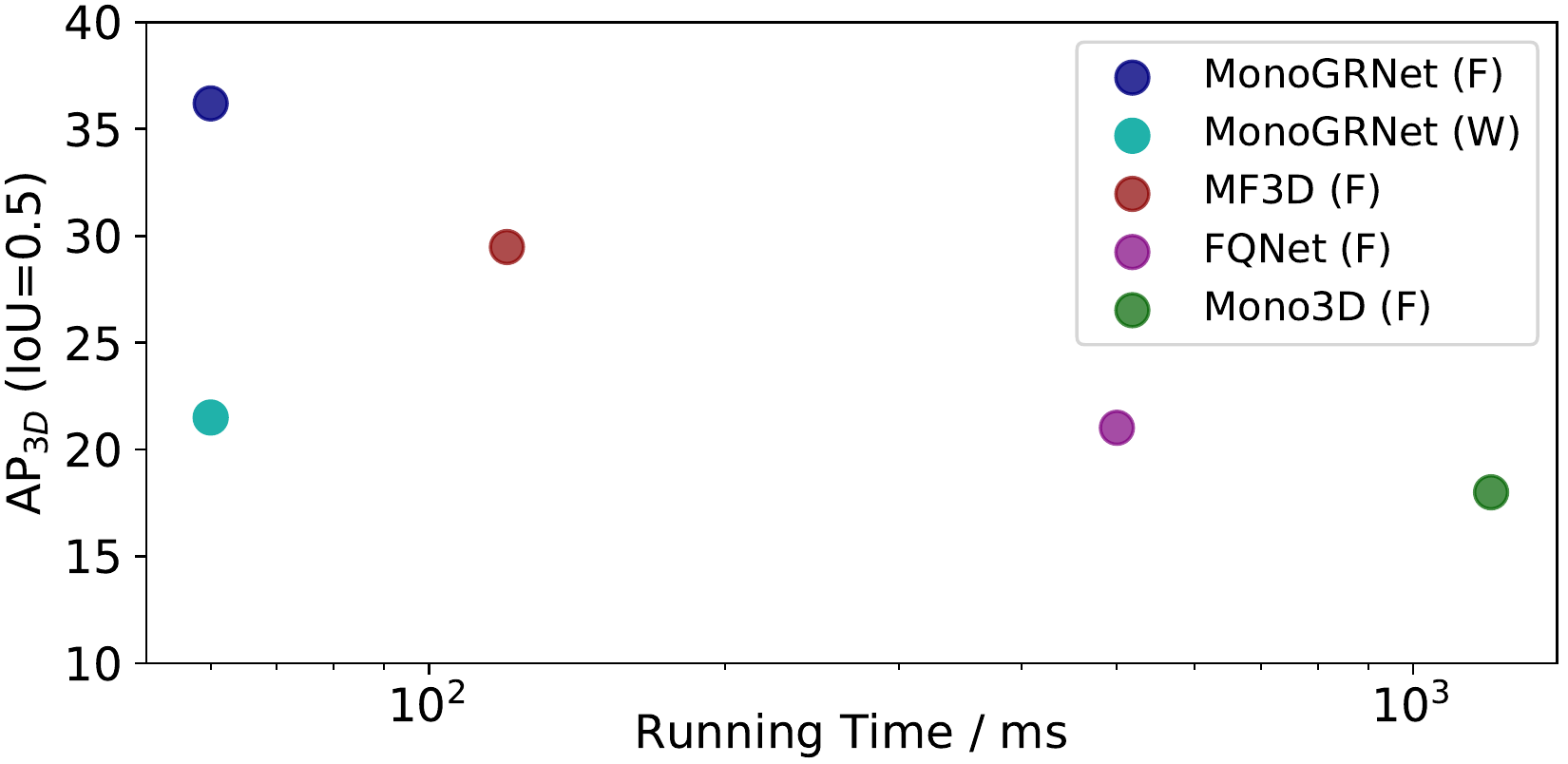}
    \caption{\textbf{Efficiency comparison.} The inference time of our method MonoGRNet (F/W) is 0.06s on a single Geforce GTX Titan X GPU on KITTI~\cite{geiger2012kitti} dataset. F and W denote fully and weakly supervised respectively.}
    \label{fig:time_ap}
\end{figure}

\begin{figure*}[!htb]
    \centering
    \includegraphics[width=1\linewidth]{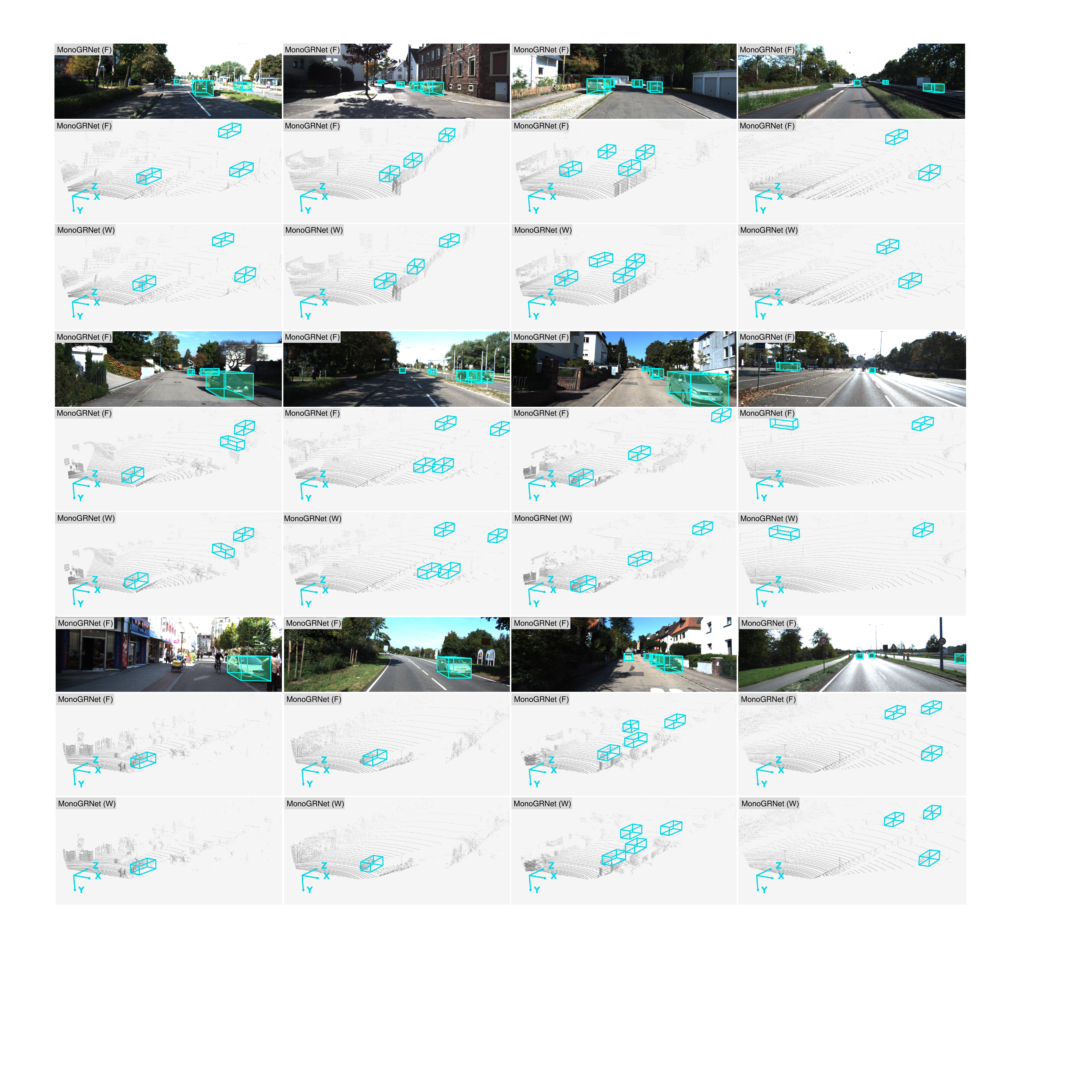}
    \caption{\textbf{Qualitative results on KITTI.} F and W are short for full and weak supervision. The predicted 3D bounding boxes are shown in images and in the 3D space from an oblique view. The proposed method has satisfactory performance even when the object is far away, occluded, in the shadows or exposed in strong light. 3D point clouds (in gray) are only for referenced visualization.}
    \label{fig:vis_det_kitti}
\end{figure*}

\begin{figure*}[!htb]
    \centering
    \includegraphics[width=0.98\linewidth]{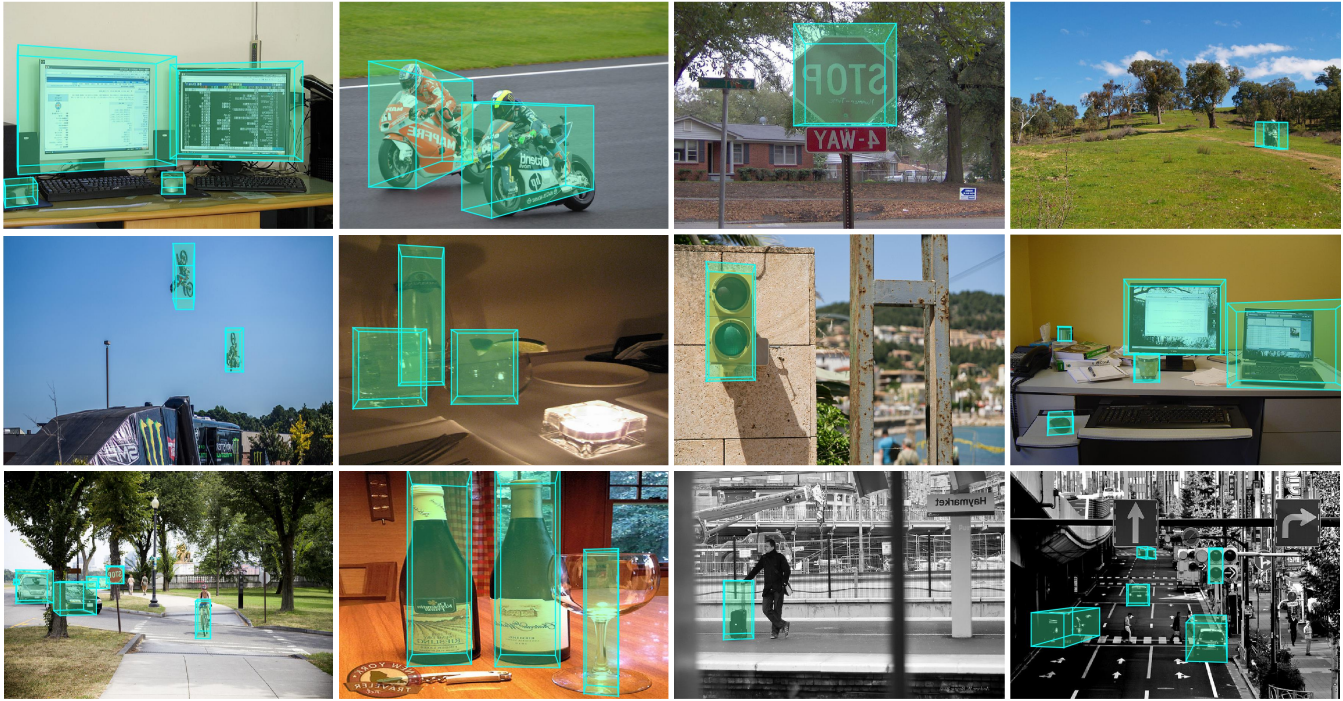}
    \caption{\textbf{Qualitative results on MS COCO.} 
    Our approach demonstrates potentials in general 3D object detection from a monocular image. We use our weakly supervised MonoGRNet (W) on MS COCO, which does not provide ground truth 3D bounding boxes to train the fully supervised MonoGRNet (F).
    }
    \label{fig:vis_det_coco}
\end{figure*}

\begin{figure*}[!htb]
	\centering
	\scriptsize
\includegraphics[width=1\linewidth]{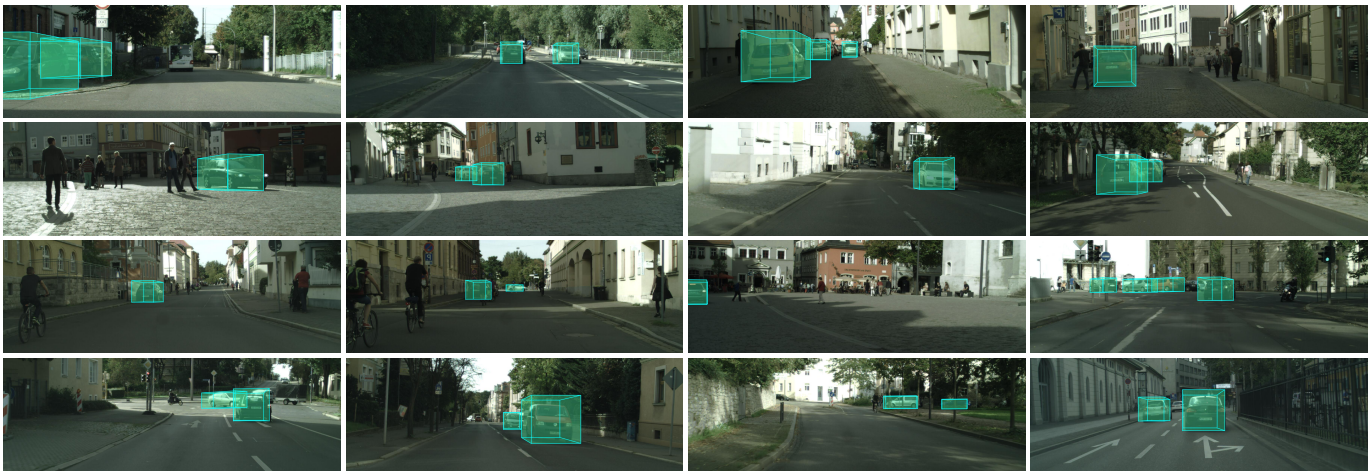}
	\caption{\textbf{Qualitative results on Cityscapes.} The experiment was conducted using MonoGRNet (W).
}
	\label{fig:vis_det_cityscape}
\end{figure*}

Different from most of the previous studies on 3D object detection, our evaluation is not limited to datasets that offer ground truth 3D bounding boxes for training. In addition to the popular KITTI~\cite{geiger2012kitti} dataset, we also experiment on the challenging Cityscapes~\cite{Cordts2016Cityscapes} and MS COCO~\cite{Lin2014Microsoft} dataset where the 3D bounding boxes are not provided. We evaluate both our fully supervised MonoGRNet (F) and weakly supervised version MonoGRNet (W).

\subsection{Implementation.}
The whole framework is implemented using Python~\cite{van1995python} and Tensorflow~\cite{tensorflow2015-whitepaper}. We employ VGG16~\cite{matthew2014vgg} pretrained on ImageNet~\cite{imagenet15} as the backbone network in MonoGRNet. We remove the original fully connected layers in VGG16~\cite{matthew2014vgg} to obtain a fully convolutional backbone. For hyperparameters, we choose $S_u \times S_v = 39 \times 12$, which is the size of the grid $\mathcal{G}$, or namely, the feature map resolution of the head layers. The whole network is trained using Adam~\cite{kin2015adam} optimizer for 40 epochs with a constant learning rate of $10^{-5}$. L2 regularization is applied to model parameters at a decay weight of $5\times10^{-5}$. 

\subsection{Qualitative Results}
Previous work could train 3D object detectors on KITTI~\cite{geiger2012kitti} but not on Cityscapes~\cite{Cordts2016Cityscapes} and MS COCO~\cite{Lin2014Microsoft}, since the last two do not offer labeled 3D bounding boxes. Our approach is not limited by 3D labels and can handle all three datasets.

\subsubsection{On KITTI Dataset.} 
The detection results in various scenarios are shown in \fig{\ref{fig:vis_det_kitti}}, including the residential areas and the highways, with objects at short and long distances. The detector is robust enough to corner cases including strong light, shadows, truncation and occlusion. This robustness is crucial in practical use. Comparing to the fully supervised MonoGRNet (F), the weakly supervised MonoGRNet (W) exhibits promising qualitative performance.

\subsubsection{On MS COCO Dataset.}
MS COCO~\cite{Lin2014Microsoft} dataset contains a wide variety of objects. Both the indoor and outdoor scenes are included. Although MS COCO does not provide ground truth 3D bounding boxes, the proposed method is not subject to this limitation and can learn from the 2D bounding box labels. The dataset does not offer the camera parameters, which cannot be simply estimated given a single RGB image. But fortunately our method only need a relative focal length, and thus we use an empirical camera configuration in the experiment. We do not apply the motion consistency loss since the neighbouring frames are inaccessible. We only test MonoGRNet (W) since the ground truth is not available for training MonoGRNet (F). The visualization results are presented in \fig{\ref{fig:vis_det_coco}}. This experiment shows the potentials of our method on general 3D object detection at low cost, i.e., based on cheap cameras can be deployed in various scenarios.

\subsubsection{On Cityscapes Dataset.} 
We test our MonoGRNet (W) on Cityscapes~\cite{Cordts2016Cityscapes} dataset without using any label on Cityscapes~\cite{Cordts2016Cityscapes} in training. We first train a Faster-RCNN~\cite{ren2017faster} 2D object detector on the KITTI dataset and directly apply it to Cityscapes to obtain 2D bounding boxes, then our network learns 3D bounding boxes using the detected 2D bounding boxes using our weakly supervised method. The proposed method exhibits stable performance as is shown in \fig{\ref{fig:vis_det_cityscape}}. This experiment demonstrates that we do not necessarily require manually labeled 2D bounding boxes when transferring to new datasets. 

\begin{figure*}[!htb]
	\centering
	\scriptsize
\includegraphics[width=0.95\linewidth]{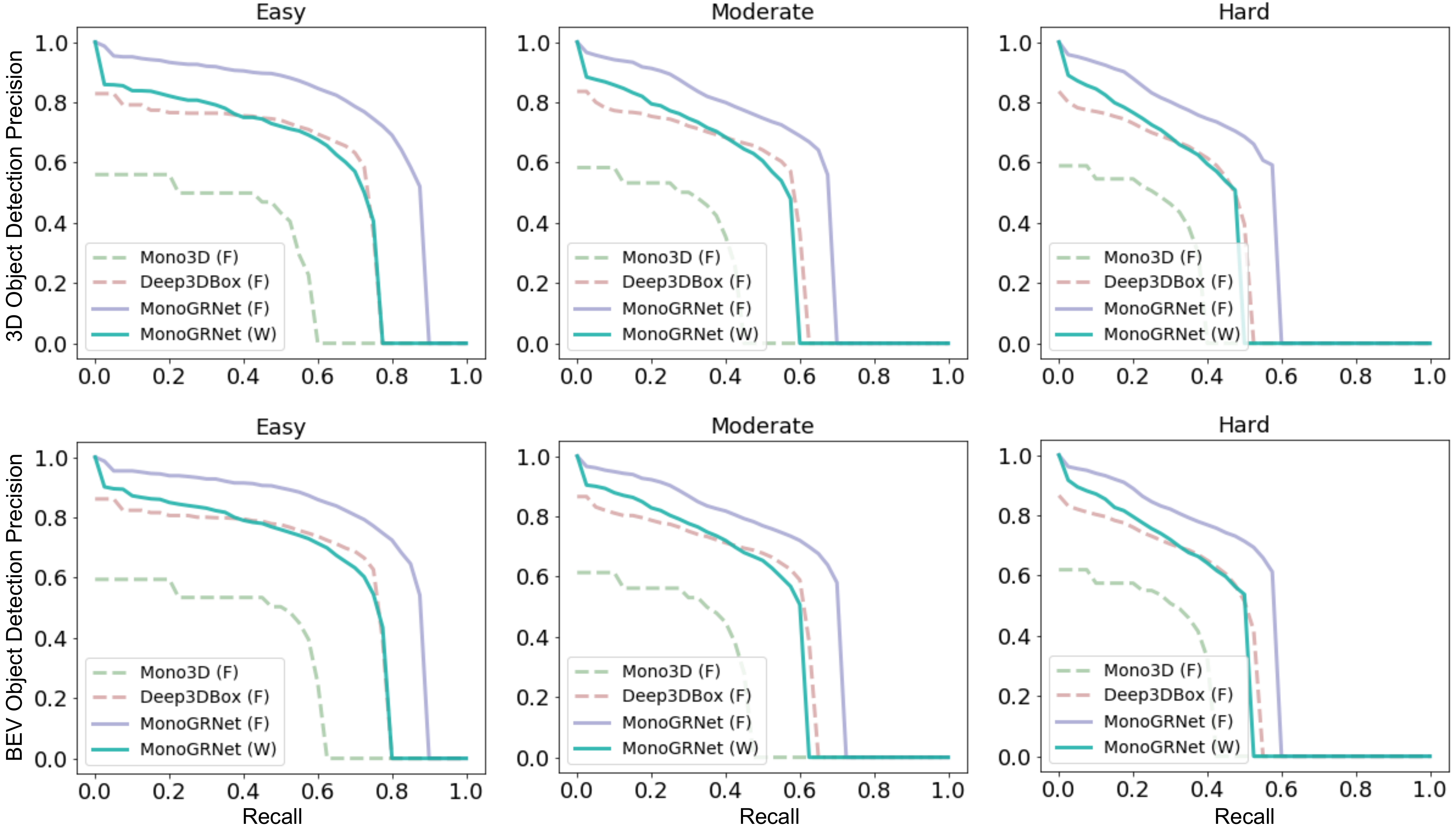}
	\caption{\textbf{Recall-precision curve of 3D and BEV object detection on KITTI \textit{val} set.} F and W are short for full and weak supervision. The 3D IoU threshold is 0.3. Although MonoGRNet (W) is learned from the weak supervision of 2D bounding boxes, it exhibits promising performance compared to methods with full supervision from 3D bounding boxes.
}
	\label{fig:pr_curve_3d_bev}
\end{figure*}

\begin{figure*}[!htb]
	\centering
	\scriptsize
\includegraphics[width=0.95\linewidth]{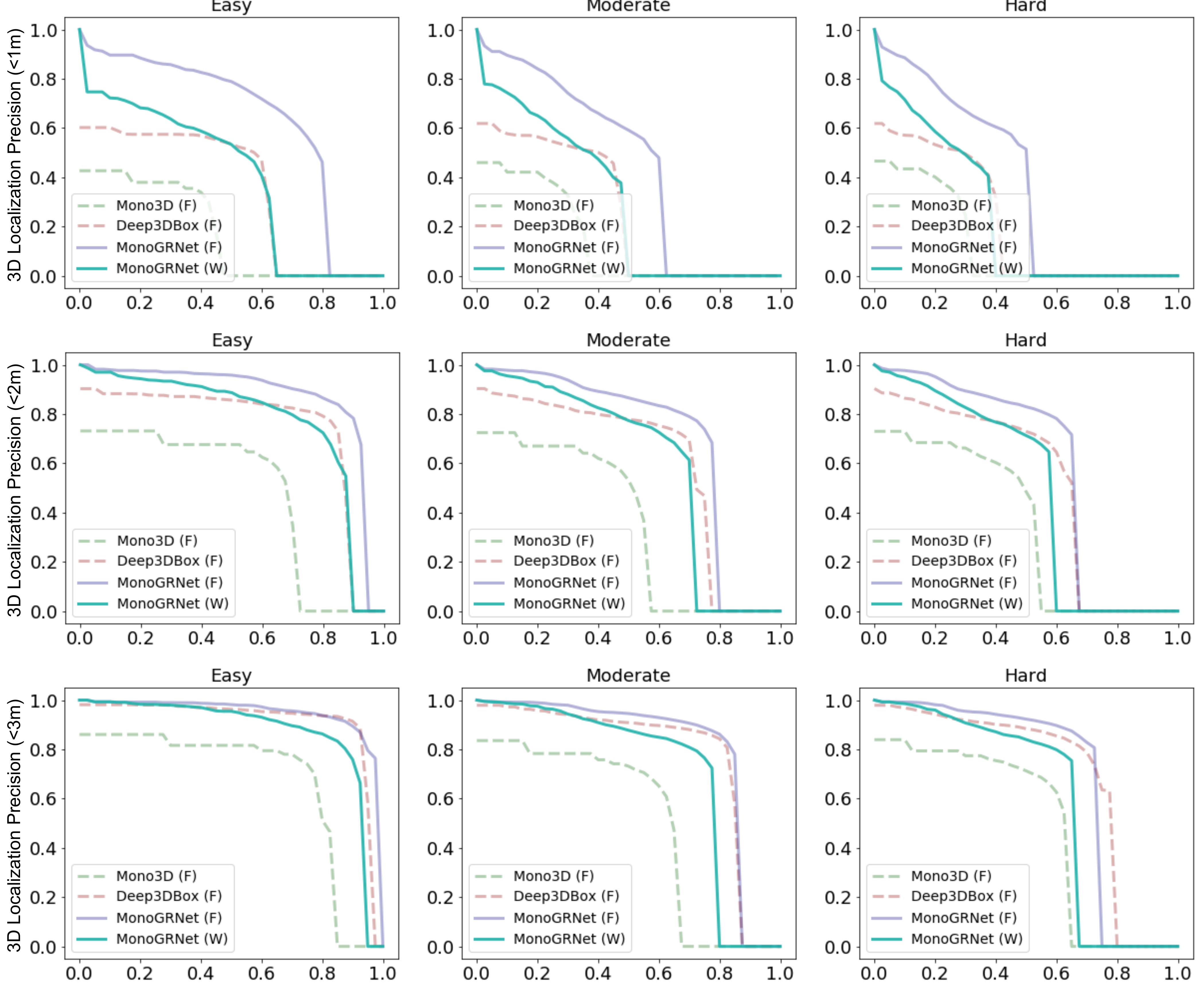}
	\caption{\textbf{Recall-precision curve of 3D object localization on KITTI \textit{val} set.} F and W are short for full and weak supervision. The distance thresholds are set to 1m, 2m and 3m, corresponding to the first, second and the third row.
}
	\label{fig:pr_curve_loc}
\end{figure*}

\begin{figure*}[!htb]
	\centering
	\scriptsize
\includegraphics[width=0.95\linewidth]{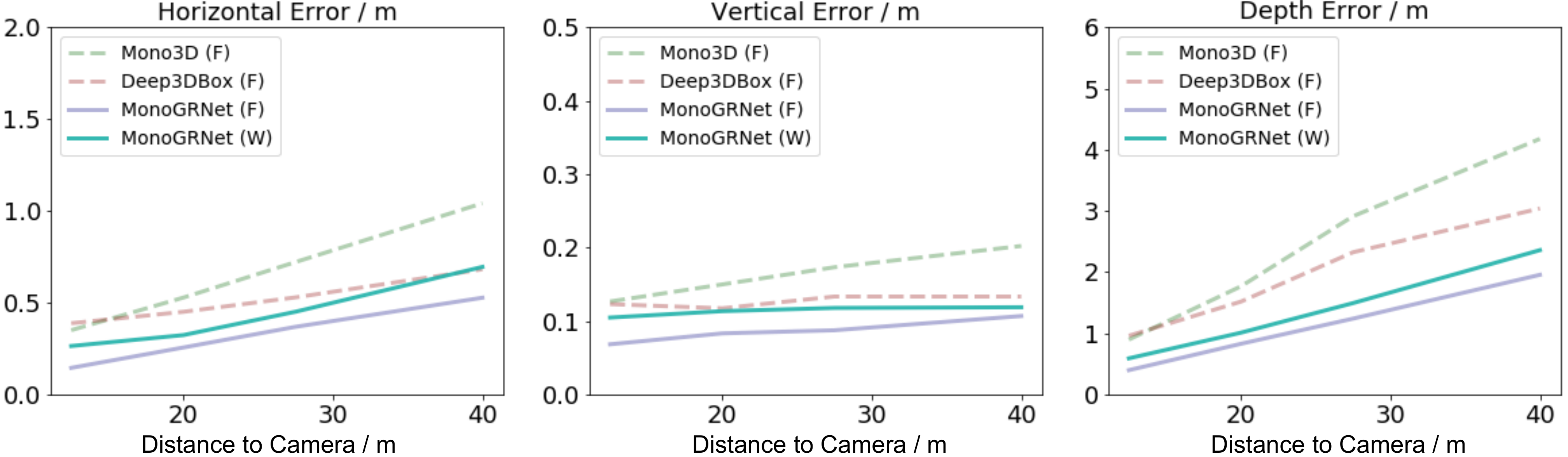}
	\caption{\textbf{3D localization error with respect to the ground truth distance on KITTI \textit{val} set.} F and W are short for full and weak supervision. MonoGRNet (F/W) demonstrates a robust localization performance.
}
	\label{fig:curve_err}
\end{figure*}

\subsection{Quantitative Results}
The quantitative experiments are conducted on the KITTI dataset but not on the Cityscapes and MS COCO where the ground truth 3D bounding boxes are not provided for quantitative evaluation. The evaluation criteria include the commonly used 3D average precision $AP_{3D}$, the bird's eye view (BEV) average precision $AP_{BEV}$ and the average orientation similarity $AOS$. $AP_{3D}$ measures the intersection of union (IoU) between the ground truth 3D bounding box and the predicted one. If the IoU is greater than a given threshold, the ground truth is successfully recalled. A higher IoU threshold means the ground truth is more difficult to recall. Different from $AP_{3D}$, $AP_{BEV}$ ignores the vertical dimension, measuring the IoU from the bird's eye view. $AOS$ measures the average orientation similarity between the predicted 3D bounding boxes and the ground truth. The calculation of $AOS$ is similar to that of average precision and more details can be found in Section~2.5 of \cite{geiger2012kitti}. All the three criteria have been widely accepted~\cite{chen2017multiview, chen20153dop} in the assessment of 3D object detectors. We follow the publicly available train-val split~\cite{chen2017multiview, chen20153dop} on KITTI~\cite{geiger2012kitti} 3D object detection dataset.

\subsubsection{Comparison to Fully Supervised Methods}
To the best of our knowledge, this is a pioneering work unifying the fully and weakly supervised learning of monocular 3D object detection. Most of the published work~\cite{cvpr16chen, mousavian20173dbox,qin2019monogr} only use fully annotated 3D bounding boxes as supervision. \fig{\ref{fig:time_ap}} compares the speed and average precision, showing that MonoGRNet (F) requires the least computational time while achieving promising average precision. The weakly supervised MonoGRNet (W) even demonstrates similar average precision with the recent full supervised method FQNet~\cite{Liu_2019_CVPR}. \fig{\ref{fig:pr_curve_3d_bev}} and \tab{\ref{tab:expap3d}} and \tab{\ref{tab:expapbev}} compare the $AP_{3D}$ and $AP_{BEV}$. \fig{\ref{fig:curve_err}} illustrates the localization error with respect to distances. It is shown that we have promising performance compared to these approaches. Even though there is still a gap between the weakly and fully supervised methods, we believe the potentially large amount of weakly labeled data can further narrow the performance gap. In addition, 3D labels for large-scale real datasets are difficult to obtain in practice, and our method has the potentials to be an alternative approach to saving the annotation cost and promote the feasibility of monocular 3D object detection in future applications.

In \tab{\ref{tab:compare_to_lidar}}, we also compare to the most recent state-of-the-art fully supervised methods~\cite{wang2019pseudo, ku2019monocular, ma2019accurate}. Although these methods demonstrate better $AP_{3D}$ than the fully supervised MonoGRNet, our results are noteworthy due to three facts. \textbf{First}, these methods~\cite{wang2019pseudo, ku2019monocular, ma2019accurate} on monocular 3D object detection utilize the 3D LiDAR point clouds as a powerful supervision to learn dense 3D reconstructions that improve the detection performance, while MonoGRNet targets at a more general scenario where we do not assume the existence of point clouds or dense depth data in any part of the training phase. In fact, a very large proportion of monocular images in the real world are acquired without 3D point clouds.
\textbf{Second}, the computational cost and time consumption of these methods~\cite{wang2019pseudo, ku2019monocular, ma2019accurate} are much more than our MonoGRNet. It is reported that the 3D reconstruction module~\cite{fu2018ordinal} used by \cite{ma2019accurate} takes $\sim$500 ms for each frame during inference, but MonoGRNet only requires $\sim$60ms under the same GPU configuration. \textbf{Third}, one of the main contributions of this paper is that we unified fully supervised and weakly supervised monocular object detection in a single framework, which can flexibly adapt to various scenarios no matter whether the fully annotated 3D bounding boxes are available in training. To the best of our knowledge, this work is the first that achieves this.

\begin{table*}[t] 
\centering
\caption{\textbf{Average precision of 3D object detection on KITTI \textit{val} set compared to previous fully supervised methods.} A ground truth object is successfully recalled if the 3D intersection of union (IoU) between its true 3D bounding box and the predicted box is no less than a certain threshold. 
Recalling an object in 3D is much more difficult than in 2D image given the extensive searching space and the lack of 3D data.
}
\begin{spacing}{1.3}
\scalebox{0.8}{
\begin{tabular}{|c|c||ccc|ccc|ccc|ccc|ccc|}
\hline
\multirow{2}{*}{Method} &   \multirow{2}{*}{SP}& \multicolumn{3}{c|}{AP\textsubscript{3D} (IoU=0.1)}  & \multicolumn{3}{c|}{AP\textsubscript{3D} (IoU=0.2)}& \multicolumn{3}{c|}{AP\textsubscript{3D} (IoU=0.3)} & \multicolumn{3}{c|}{AP\textsubscript{3D} (IoU=0.5)}  & \multicolumn{3}{c|}{AP\textsubscript{3D} (IoU=0.7)} \\ \cline{3-17} 
					    &                         &\multicolumn{1}{c|}{Easy} & \multicolumn{1}{c|}{Mod.} & Hard &\multicolumn{1}{c|}{Easy} & \multicolumn{1}{c|}{Mod.} & Hard & \multicolumn{1}{c|}{Easy}  & \multicolumn{1}{c|}{Mod.} & Hard  & \multicolumn{1}{c|}{Easy} & \multicolumn{1}{c|}{Mod.} & Hard & \multicolumn{1}{c|}{Easy} & \multicolumn{1}{c|}{Mod.} & Hard \\ \hline

Mono3D~\cite{cvpr16chen}     &Full 	& 54.71 & 43.67  & 39.59 & 41.21 & 33.40 & 28.89 & 28.29	& 23.21	& 19.49 & 25.19 & 18.20 & 15.22  & 2.53 & 2.31 & 2.31\\

Deep3DBox~\cite{mousavian20173dbox}  &Full & 82.57 & 69.74 & 60.86 & 69.97 & 56.62 & 48.59 & 54.30 & 43.42 & 36.57 & 27.04 & 20.55 & 15.88 & 5.85 & 4.10 & 3.84 \\

MF3D~\cite{xu2018multifusion}  &Full & -	& -	& - & - & - & - & - & - & - & \textbf{47.88} & 29.48 & 26.44 & 10.53 & 5.69 &5.39 \\

FQNet~\cite{Liu_2019_CVPR} &Full & -	& -	& - & - & - & - & - & - & - & 28.16 & 21.02& 19.91 & 5.98 & 5.50 & 4.75 \\

ROI-10D~\cite{manhardt2019roi} &Full & -	& -	& - & - & - & - & - & - & - & 37.59 & 25.14 & 21.83 & 6.91 & 6.63 & 6.29\\

MonoGRNet (F)     &Full & \textbf{88.10} & \textbf{76.89} & \textbf{67.76} & \textbf{82.18} & \textbf{64.54} & \textbf{55.63} & \textbf{73.66} & \textbf{61.74} & \textbf{47.75} & 43.66 & \textbf{36.20} & \textbf{30.22} & \textbf{13.88} & \textbf{10.19} & \textbf{7.62}\\

MonoGRNet (W)                           &Weak  & 80.70 & 64.42 & 55.42 & 69.60 & 53.70 & 45.45 & 56.16 & 42.61 & 35.36 & 25.66 & 21.57 & 17.40  & 6.92 & 5.63 & 4.89 \\

\hline
\end{tabular}
}
\end{spacing}
\label{tab:expap3d}
\end{table*}

\begin{table*}[t] 
\centering
\caption{\textbf{Average precision of bird's eye view (BEV) object detection on KITTI \textit{val} set compared to previous fully supervised methods.} A ground truth object is successfully recalled if the BEV intersection of union (IoU) between its true 3D bounding box and the predicted box is no less than a certain threshold.}
\begin{spacing}{1.3}
\scalebox{0.8}{
\begin{tabular}{|c|c||ccc|ccc|ccc|ccc|ccc|}
\hline
\multirow{2}{*}{Method} &   \multirow{2}{*}{SP}& \multicolumn{3}{c|}{AP\textsubscript{BEV} (IoU=0.1)}  & \multicolumn{3}{c|}{AP\textsubscript{BEV} (IoU=0.2)}  & \multicolumn{3}{c|}{AP\textsubscript{BEV}  (IoU=0.3)} & \multicolumn{3}{c|}{AP\textsubscript{BEV} ((IoU=0.5)}  & \multicolumn{3}{c|}{AP\textsubscript{BEV} ((IoU=0.7)} \\ \cline{3-17} 
					    &                         &\multicolumn{1}{c|}{Easy} & \multicolumn{1}{c|}{Mod.} & Hard &\multicolumn{1}{c|}{Easy} & \multicolumn{1}{c|}{Mod.} & Hard & \multicolumn{1}{c|}{Easy}  & \multicolumn{1}{c|}{Mod.} & Hard  & \multicolumn{1}{c|}{Easy} & \multicolumn{1}{c|}{Mod.} & Hard & \multicolumn{1}{c|}{Easy} & \multicolumn{1}{c|}{Mod.} & Hard \\ \hline

Mono3D~\cite{cvpr16chen}     &Full 	& 55.47 & 44.25   & 40.14 & 46.26 & 35.22 &  34.23 & 32.76	& 25.15	& 23.65	& 30.50	& 22.39	& 19.16 & 5.22 & 5.19 & 4.13 \\
Deep3DBox~\cite{mousavian20173dbox}  &Full & 84.29	& 70.40	& 61.49 & 71.27 &  58.88 & 49.92 & 57.14	& 47.20	& 39.06	& 30.17	& 23.77	& 18.84 & 9.99 & 7.71 & 5.30\\

MF3D~\cite{xu2018multifusion}  &Full & -	& -	& - & - & - & - & - & - & - &  \textbf{55.02}	& 36.73	& 31.27 & 22.03 & 13.63 & 11.60\\

FQNet~\cite{Liu_2019_CVPR} &Full & -	& -	& - & - & - & - & - & - & - & 32.57	&  24.60 & 21.25 & 9.50 & 8.02 & 7.71 \\

ROI-10D~\cite{manhardt2019roi} &Full & -	& -	& - & - & - & - & - & - & - & 46.85 & 34.05 & 30.46 & 14.50 & 9.91 & 8.73\\

MonoGRNet (F)     &Full & \textbf{88.22} & \textbf{77.09}	& \textbf{67.91} & \textbf{83.27} &\textbf{65.13} & \textbf{56.18} & \textbf{74.92}	& \textbf{63.22}	&  \textbf{54.65}	& 52.52	& \textbf{39.98}	& \textbf{33.14} & \textbf{24.97} & \textbf{19.44} & \textbf{16.30} \\

MonoGRNet (W)                           &Weak  & 81.12     & 64.76	& 55.76 & 71.38 &  60.11 & 46.27 & 58.61 & 48.75	& 41.49 & 32.23 & 26.88	& 22.47 & 12.54 & 9.67 & 8.25\\

\hline
\end{tabular}
}
\end{spacing}
\label{tab:expapbev}
\end{table*}

\begin{table*}[!htb]
\centering
\caption{\textbf{Average precision of 3D and bird's eye view (BEV) object detection on KITTI \textit{val} set compared to weakly supervised baselines.} BA and OC are short for background-aware and object-centric respectively. MinProjErr denotes the baseline method derived from \cite{mousavian20173dbox} to recover 3D location as described in \sect{\ref{sect:qual_result}}. GeoGL indicates the proposed geometry-guided learning of 3D location.
}
\setlength{\tabcolsep}{4mm}{
\begin{spacing}{1.3}
\scalebox{0.815}{
\begin{tabular}{|cc||ccc|ccc|}
\hline
\multicolumn{2}{|c||}{Method}                        & \multicolumn{3}{c|}{AP\textsubscript{3D} / AP\textsubscript{BEV}  (IoU=0.3)} & \multicolumn{3}{c|}{AP\textsubscript{3D} / AP\textsubscript{BEV} (IoU=0.5)}  \\ \hline
\multicolumn{1}{|c|}{Orientation} & 3D Location  & \multicolumn{1}{c|}{Easy}  & \multicolumn{1}{c|}{Moderate} & Hard  & \multicolumn{1}{c|}{Easy} & \multicolumn{1}{c|}{Moderate} & Hard \\ \hline

$\text{BA}\times\text{2}$   & MinProjErr             &  37.67  /  42.80 &  28.50 / 35.69  & 23.60 /  29.99 & 15.38 / 18.63 & 13.19 / 15.70 &  11.41 / 14.80 \\
$\text{BA}\times\text{1}$   & MinProjErr             &  39.30  /  43.92 &  29.80 / 36.39  & 24.40 /  30.53 & 16.81 / 21.47 & 14.10 / 18.38 &  11.78 / 15.16 \\
$\text{BA}\times\text{2}$   & GeoGL~(Ours)   &  50.10  /  54.31 &  39.16 / 41.24  & 32.57 /  34.35 & 22.35 / 28.13 & 18.72 / 21.65 &  15.85 / 17.45 \\
$\text{BA}\times\text{1}$   & GeoGL~(Ours)   &  54.44  /  57.36 &  41.40 / 47.59  & 34.39 /  35.77 & 25.03 / 31.15 & 20.85 / 25.97 &  17.10 / 21.37 \\
OC~(Ours)                        & GeoGL~(Ours)   &  56.16  /  58.61 &  42.61 / 48.75  & 35.36 /  41.49 & 25.66 / 32.23 & 21.57 / 26.88 &  17.40 / 22.47 \\

\hline
\end{tabular}}
\end{spacing}}

\label{tab:3dbevap}
\end{table*}

\begin{table}[!htb]
\centering
\caption{\textbf{Comparison to fully supervised methods involving LiDAR or only monocular information during training.} The numbers are reported on KITTI \textit{val} set.}
\setlength{\tabcolsep}{2.2mm}{
\begin{spacing}{1.3}
\scalebox{0.72}{
\begin{tabular}{|c|r|cc||ccc|}
\hline
\multirow{2}{*}{Method} &  \multirow{2}{*}{Time~} & \multicolumn{2}{c||}{Data}  & \multicolumn{3}{c|}{AP\textsubscript{3D}~(IoU=0.5)}  \\ \cline{3-7} 
					    &                          &\multicolumn{1}{c|}{Train} & \multicolumn{1}{c||}{Eval} & \multicolumn{1}{c|}{Easy}  & \multicolumn{1}{c|}{Mod.} & Hard  \\ \hline
P-LiDAR~\cite{wang2019pseudo}    & $>$ 0.5 s      & LiDAR+Mono & Mono & 66.3 & 42.3 & 38.5 \\
MonoPSR~\cite{ku2019monocular}   & 0.2 s          & LiDAR+Mono & Mono & 49.65 & 41.71 & 29.95 \\
Ma \textit{et al.}~\cite{ma2019accurate} & $>$ 0.5 s & LiDAR+Mono & Mono & 68.86 & 49.19 & 42.24 \\
FQNet~\cite{Liu_2019_CVPR} & $>$ 0.4 s              & Mono         & Mono   & 28.16 & 21.02 & 19.91 \\
MonoGRNet (F)              & 0.06 s                & Mono          & Mono   & 43.66 & 36.20 & 30.22 \\
\hline
\end{tabular}
}
\end{spacing}}
\label{tab:compare_to_lidar}
\end{table}

\begin{table}[!htb]
\centering
\caption{\textbf{Average orientation similarity on KITTI \textit{val} set.} The three approaches use GeoGL to predict the 3D location. The only difference is how they learn object orientations.}
\setlength{\tabcolsep}{2.8mm}{
\begin{spacing}{1.3}
\scalebox{0.75}{
\begin{tabular}{|c||ccc|ccc|}
\hline
\multirow{2}{*}{Method} &  \multicolumn{3}{c|}{AOS~(IoU=0.3)}  & \multicolumn{3}{c|}{AOS~(IoU=0.5)}  \\ \cline{2-7} 
					    &\multicolumn{1}{c|}{Easy} & \multicolumn{1}{c|}{Moderate} & Hard & \multicolumn{1}{c|}{Easy}  & \multicolumn{1}{c|}{Moderate} & Hard  \\ \hline

$\text{BA}\times 2$  & 64.23   & 62.86 & 49.96  & 64.14  & 56.93   & 49.85   \\
$\text{BA}\times 1$  & 83.58   & 80.54 & 64.13   & 83.56  & 73.09  &  63.97  \\
OC~(Ours)            & 90.00   & 88.78 & 70.89  & 89.94  &  80.31 & 70.76  \\
\hline
\end{tabular}
}
\end{spacing}}

\label{tab:orientation_similarity} 
\end{table}

\begin{table}[h]
\centering
\caption{\textbf{Ablation study on loss functions for weakly supervised method on KITTI \textit{val} set.} The loss functions are crucial in learning the 3D location of objects using their ground truth 2D bounding boxes and the motion consistency across unlabeled frames}
 \begin{spacing}{1.1}
\scalebox{0.9}{
\setlength{\tabcolsep}{2.6mm}{
\begin{tabular}{|cccc||ccc|}
\hline
   \multicolumn{4}{|c||}{Loss Configuration}  & \multicolumn{3}{c|}{$AP_{3D}$~(IoU=0.3)} \\ \hline
   
   $\mathcal{L}_{Z_c}$ & $\mathcal{L}_{\mathbf{c}}$ & $\mathcal{L}_{\Delta\mathbf{C}}$ & $\mathcal{L}_{a}$ & \multicolumn{1}{c|}{Easy} & \multicolumn{1}{c|}{Moderate} & Hard \\ \hline
                &            &            &            & 38.28 & 28.56 & 24.13 \\
     \checkmark &            &            &            & 44.50 & 32.33 & 29.10  \\
    \checkmark  & \checkmark &            &            & 52.72 & 40.59 & 33.71 \\
    \checkmark  & \checkmark & \checkmark &            & 54.21 & 41.19 & 34.19 \\
    \checkmark  & \checkmark & \checkmark & \checkmark & 56.16 & 42.61 & 35.36 \\ \hline
\end{tabular}
}}
\end{spacing}

\label{tab:ablation}
\end{table}

\subsubsection{Comparison to Weakly Supervised Baselines}
\label{sect:qual_result}
We also examine our approach by comparing to weakly supervised baselines. As no weakly supervised monocular 3D object detector has been published, we borrowed ideas from previous transfer learning methods~\cite{torrey2010transfer} and state-of-the-art monocular 3D object detectors~\cite{mousavian20173dbox, qin2019monogr} to construct baseline approaches that only require annotated 2D bounding boxes in training. In our object-centric transfer learning of orientation, the background is mostly ignored, and the teacher network mainly focuses on the object itself. Previous work~\cite{torrey2010transfer} on transfer learning usually takes the whole image as input, making image-level transfer learning. Other work~\cite{Chen20173D} on 3D object detection also leverages the background information around the objects to improve the detection performance. In the baseline methods, we enlarge the 2D bounding boxes before cropping the image as is shown in \fig{\ref{fig:trans_learn}}, allowing more background into the crop. The baselines for orientation prediction are named $\text{Background-Aware~(BA)}\times\lambda$, where the height and width of the 2D bounding boxes are both enlarged $\lambda + 1$ times, i.e., the original $B_{2D}=(w_{2D}, h_{2D}, b_u, b_v)$ is expanded to $B_{2D}=\left[(\lambda+1)w_{2D}, (\lambda + 1) h_{2D}, b_u, b_v\right]$. $B_{2D}$ is truncated by the image boundaries. As $\lambda$ becomes larger, the baseline is increasingly similar to methods feeding the whole image as input. 

In our geometry-guided learning of 3D location, the network learns from the 3D-to-2D projective geometric constrain and the motion consistency. Inspired by Deep3DBox~\cite{mousavian20173dbox}, we develop another baseline method to obtain the 3D location from the 2D bounding box. First, we detect the 2D bounding boxes on image. Then, we find the corresponding 3D bounding box locations by minimizing their projection error with the associated 2D boxes. This baseline approach to recovering the 3D location is denoted as MinProjErr in our experiments. Results are shown in \tab{\ref{tab:3dbevap}} and \tab{\ref{tab:orientation_similarity}}. It is clear that our geometry-guided learning method can bring significant performance improvement in all the listed criteria. The object-centric transfer learning performs better than the background-aware counterparts.

Recently, the work~\cite{guizilini20203d, godard2019digging} proposed self-supervised methods to learn a dense monocular depth predictor. The absolute relative error (AbsRel) achieved by these methods are 0.07m for~\cite{guizilini20203d} and 0.11m for~\cite{godard2019digging}, while the AbsRel of our weakly supervised instance-level depth estimation is only 0.05m. Besides, the dense pixel-level depth predicted by these methods is the the depth of object surfaces, while the instance depth predicted by our MonoGRNet is the 3D center depth of an object. The latter can be directed used to compute the 3D center coordinates of the targeted objects via \eqn{\ref{eq:2d_3d}}, but the former requires additional steps that could bring extra computational cost.

\subsubsection{Ablation Study on Loss Functions}
The loss functions proposed in \sect{\ref{sect:learning_weak_loc}} are crucial in supervising the 3D location using labeled 2D bounding boxes. In order to examine the effectiveness of the loss functions, we experiment with different loss configurations and present the results in Table~\ref{tab:ablation}. If none of $\mathcal{L}_{Z_c}$, $\mathcal{L}_{\mathbf{c}}$ or $\mathcal{L}_{\Delta\mathbf{C}}$ is applied, we obtain $\mathbf{C}$ by minimizing the re-projection error. It is shown that $\mathcal{L}_{Z_c}$ and $\mathcal{L}_{\mathbf{c}}$ can bring a considerable performance gain, while $\mathcal{L}_{\Delta\mathbf{C}}$ can further refine the prediction.

\section{Conclusion}

We have presented the MonoGRNet framework for 3D object detection from monocular images. MonoGRNet decomposes the monocular 3D object detection task into four sub-tasks, which are 2D object detection, instance-level depth estimation, projected 3D center estimation and local corner regression. The task decomposition allows the 3D bounding boxes to be predicted in a single forward pass without the object-proposal stage or the computationally expensive pixel-level depth prediction. We also demonstrate that the framework can flexibly adapt to both fully and weakly supervised learning without changing the network structure or most of the loss functions. Extensive experiments are conducted on three public datasets, KITTI, Cityscapes and MS COCO. Although the last two does not offer ground truth 3D bounding boxes, our framework is still trainable in such a setting. Qualitative and quantitative results have shown the promising performance of our method in diverse scenarios.

\bibliographystyle{IEEEtran}
\bibliography{reference}

\begin{thebibliography}{10}
\providecommand{\url}[1]{#1}
\csname url@samestyle\endcsname
\providecommand{\newblock}{\relax}
\providecommand{\bibinfo}[2]{#2}
\providecommand{\BIBentrySTDinterwordspacing}{\spaceskip=0pt\relax}
\providecommand{\BIBentryALTinterwordstretchfactor}{4}
\providecommand{\BIBentryALTinterwordspacing}{\spaceskip=\fontdimen2\font plus
\BIBentryALTinterwordstretchfactor\fontdimen3\font minus
  \fontdimen4\font\relax}
\providecommand{\BIBforeignlanguage}[2]{{%
\expandafter\ifx\csname l@#1\endcsname\relax
\typeout{** WARNING: IEEEtran.bst: No hyphenation pattern has been}%
\typeout{** loaded for the language `#1'. Using the pattern for}%
\typeout{** the default language instead.}%
\else
\language=\csname l@#1\endcsname
\fi
#2}}
\providecommand{\BIBdecl}{\relax}
\BIBdecl

\bibitem{yang2018pixor}
B.~Yang, W.~Luo, and R.~Urtasun, ``Pixor: Real-time 3d object detection from
  point clouds,'' in \emph{Proceedings of the IEEE conference on Computer
  Vision and Pattern Recognition}, 2018, pp. 7652--7660.

\bibitem{zhou2018voxelnet}
Y.~Zhou and O.~Tuzel, ``Voxelnet: End-to-end learning for point cloud based 3d
  object detection,'' in \emph{Proceedings of the IEEE Conference on Computer
  Vision and Pattern Recognition}, 2018, pp. 4490--4499.

\bibitem{lang2019pointpillars}
A.~H. Lang, S.~Vora, H.~Caesar, L.~Zhou, J.~Yang, and O.~Beijbom,
  ``Pointpillars: Fast encoders for object detection from point clouds,'' in
  \emph{Proceedings of the IEEE Conference on Computer Vision and Pattern
  Recognition}, 2019, pp. 12\,697--12\,705.

\bibitem{chen2017multiview}
X.~Chen, H.~Ma, J.~Wan, B.~Li, and T.~Xia, ``Multi-view 3d object detection
  network for autonomous driving,'' in \emph{IEEE CVPR}, vol.~1, no.~2, 2017,
  p.~3.

\bibitem{qi2017frustum}
C.~R. Qi, W.~Liu, C.~Wu, H.~Su, and L.~J. Guibas, ``Frustum pointnets for 3d
  object detection from rgb-d data,'' \emph{arXiv preprint arXiv:1711.08488},
  2017.

\bibitem{hu2018joint}
H.-N. Hu, Q.-Z. Cai, D.~Wang, J.~Lin, M.~Sun, P.~Kr{\"a}henb{\"u}hl,
  T.~Darrell, and F.~Yu, ``Joint monocular 3d vehicle detection and tracking,''
  \emph{arXiv preprint arXiv:1811.10742}, 2018.

\bibitem{qin2019monogr}
Z.~Qin, J.~Wang, and Y.~Lu, ``Monogrnet: {A} geometric reasoning network for
  monocular 3d object localization,'' \emph{AAAI}, 2019.

\bibitem{chabot2017deepmanta}
F.~Chabot, M.~Chaouch, J.~Rabarisoa, C.~Teuli{\`e}re, and T.~Chateau, ``Deep
  manta: A coarse-to-fine many-task network for joint 2d and 3d vehicle
  analysis from monocular image,'' in \emph{Computer Vision and Pattern
  Recognit.(CVPR)}, 2017, pp. 2040--2049.

\bibitem{mousavian20173dbox}
A.~Mousavian, D.~Anguelov, J.~Flynn, and J.~Ko{\v{s}}eck{\'a}, ``3d bounding
  box estimation using deep learning and geometry,'' in \emph{IEEE Conference
  on Computer Vision and Pattern Recognition (CVPR)}.\hskip 1em plus 0.5em
  minus 0.4em\relax IEEE, 2017, pp. 5632--5640.

\bibitem{You2019PseudoLiDARAD}
Y.~You, Y.~Wang, W.-L. Chao, D.~Garg, G.~Pleiss, B.~Hariharan, M.~Campbell, and
  K.~Q. Weinberger, ``Pseudo-lidar++: Accurate depth for 3d object detection in
  autonomous driving,'' \emph{ArXiv}, vol. abs/1906.06310, 2019.

\bibitem{wang2019pseudo}
Y.~Wang, W.-L. Chao, D.~Garg, B.~Hariharan, M.~Campbell, and K.~Weinberger,
  ``Pseudo-lidar from visual depth estimation: Bridging the gap in 3d object
  detection for autonomous driving,'' in \emph{CVPR}, 2019.

\bibitem{Weng2019pseudolidar}
X.~Weng and K.~M. Kitani, ``Monocular 3d object detection with pseudo-lidar
  point cloud,'' \emph{ArXiv}, vol. abs/1903.09847, 2019.

\bibitem{ding2019d4lcn}
M.~Ding, Y.~Huo, H.~Yi, Z.~Wang, J.~Shi, Z.~Lu, and P.~Luo, ``Learning
  depth-guided convolutions for monocular 3d object detection,'' \emph{arXiv
  preprint arXiv:1912.04799}, 2019.

\bibitem{Jrgensen2019iouloss}
E.~J{\"o}rgensen, C.~Zach, and F.~Kahl, ``Monocular 3d object detection and box
  fitting trained end-to-end using intersection-over-union loss,''
  \emph{ArXiv}, vol. abs/1906.08070, 2019.

\bibitem{ku2019monocular}
J.~Ku, A.~D. Pon, and S.~L. Waslander, ``Monocular 3d object detection
  leveraging accurate proposals and shape reconstruction,'' in
  \emph{Proceedings of the IEEE Conference on Computer Vision and Pattern
  Recognition}, 2019, pp. 11\,867--11\,876.

\bibitem{chen2016monocular}
X.~Chen, K.~Kundu, Z.~Zhang, H.~Ma, S.~Fidler, and R.~Urtasun, ``Monocular 3d
  object detection for autonomous driving,'' in \emph{Conference on Computer
  Vision and Pattern Recognition (CVPR)}, 2016, pp. 2147--2156.

\bibitem{he2019mono3d++}
T.~He and S.~Soatto, ``Mono3d++: Monocular 3d vehicle detection with two-scale
  3d hypotheses and task priors,'' \emph{arXiv preprint arXiv:1901.03446},
  2019.

\bibitem{Cai2020heightguided}
Y.~Cai, B.~Li, Z.~Jiao, H.~Li, X.~Zeng, and X.~Wang, ``Monocular 3d object
  detection with decoupled structured polygon estimation and height-guided
  depth estimation,'' \emph{ArXiv}, vol. abs/2002.01619, 2020.

\bibitem{Liu2020SMOKESMKeypoint}
Z.~Liu, Z.~Wu, and R.~T{\'o}th, ``Smoke: Single-stage monocular 3d object
  detection via keypoint estimation,'' \emph{ArXiv}, vol. abs/2002.10111, 2020.

\bibitem{Barabanau2019Monocular3OKeypoint}
I.~Barabanau, A.~Artemov, E.~V. Burnaev, and V.~Y. Murashkin, ``Monocular 3d
  object detection via geometric reasoning on keypoints,'' \emph{ArXiv}, vol.
  abs/1905.05618, 2019.

\bibitem{Li2020RTM3DRM}
P.-X. Li, H.~Zhao, P.~Liu, and F.~Cao, ``Rtm3d: Real-time monocular 3d
  detection from object keypoints for autonomous driving,'' \emph{ArXiv}, vol.
  abs/2001.03343, 2020.

\bibitem{papadopoulos2017extreme}
D.~P. Papadopoulos, J.~R. Uijlings, F.~Keller, and V.~Ferrari, ``Extreme
  clicking for efficient object annotation,'' in \emph{Proceedings of the IEEE
  International Conference on Computer Vision}, 2017, pp. 4930--4939.

\bibitem{xiang_wacv14}
Y.~Xiang, R.~Mottaghi, and S.~Savarese, ``Beyond pascal: A benchmark for 3d
  object detection in the wild,'' in \emph{IEEE Winter Conference on
  Applications of Computer Vision (WACV)}, 2014.

\bibitem{Chang2015ShapeNetAI}
A.~X. Chang, T.~A. Funkhouser, L.~J. Guibas, P.~Hanrahan, Q.-X. Huang, Z.~Li,
  S.~Savarese, M.~Savva, S.~Song, H.~Su, J.~Xiao, L.~Yi, and F.~Yu, ``Shapenet:
  An information-rich 3d model repository,'' \emph{ArXiv}, vol. abs/1512.03012,
  2015.

\bibitem{geiger2012kitti}
A.~Geiger, P.~Lenz, and R.~Urtasun, ``Are we ready for autonomous driving? the
  kitti vision benchmark suite,'' in \emph{Computer Vision and Pattern
  Recognition (CVPR)}.\hskip 1em plus 0.5em minus 0.4em\relax IEEE, 2012, pp.
  3354--3361.

\bibitem{Cordts2016Cityscapes}
M.~Cordts, M.~Omran, S.~Ramos, T.~Rehfeld, M.~Enzweiler, R.~Benenson,
  U.~Franke, S.~Roth, and B.~Schiele, ``The cityscapes dataset for semantic
  urban scene understanding,'' in \emph{Proc. of the IEEE Conference on
  Computer Vision and Pattern Recognition (CVPR)}, 2016.

\bibitem{Lin2014Microsoft}
T.~Y. Lin, M.~Maire, S.~Belongie, J.~Hays, P.~Perona, D.~Ramanan, P.~Dollár,
  and C.~L. Zitnick, ``Microsoft coco: Common objects in context,'' 2014.

\bibitem{girshick2015fast}
R.~Girshick, ``Fast r-cnn,'' in \emph{Proceedings of the IEEE international
  conference on computer vision}, 2015, pp. 1440--1448.

\bibitem{ren2017faster}
S.~Ren, K.~He, R.~Girshick, and J.~Sun, ``Faster r-cnn: towards real-time
  object detection with region proposal networks,'' \emph{IEEE Transactions on
  Pattern Analysis \& Machine Intelligence}, 2017.

\bibitem{redmon2016yolo}
J.~Redmon, S.~Divvala, R.~Girshick, and A.~Farhadi, ``You only look once:
  Unified, real-time object detection,'' in \emph{Proceedings of the IEEE
  conference on computer vision and pattern recognition}, 2016, pp. 779--788.

\bibitem{redmon2017yolo9000}
J.~Redmon and A.~Farhadi, ``Yolo9000: Better, faster, stronger,'' in
  \emph{Computer Vision and Pattern Recognition (CVPR)}.\hskip 1em plus 0.5em
  minus 0.4em\relax IEEE, 2017, pp. 6517--6525.

\bibitem{liu2016ssd}
W.~Liu, D.~Anguelov, D.~Erhan, C.~Szegedy, S.~Reed, C.-Y. Fu, and A.~C. Berg,
  ``Ssd: Single shot multibox detector,'' in \emph{European conference on
  computer vision (ECCV)}.\hskip 1em plus 0.5em minus 0.4em\relax Springer,
  2016, pp. 21--37.

\bibitem{fu2017dssd}
C.-Y. Fu, W.~Liu, A.~Ranga, A.~Tyagi, and A.~C. Berg, ``Dssd: Deconvolutional
  single shot detector,'' \emph{arXiv preprint arXiv:1701.06659}, 2017.

\bibitem{teichmann2016multinet}
M.~Teichmann, M.~Weber, M.~Zoellner, R.~Cipolla, and R.~Urtasun, ``Multinet:
  Real-time joint semantic reasoning for autonomous driving,'' \emph{arXiv
  preprint arXiv:1612.07695}, 2016.

\bibitem{he2017mrcn}
K.~He, G.~Gkioxari, P.~Doll\'ar, and R.~Girshick, ``Mask r-cnn,'' \emph{arXiv
  preprint arXiv:1703.06870}, 2017.

\bibitem{xu2018multifusion}
B.~Xu and Z.~Chen, ``Multi-level fusion based 3d object detection from
  monocular images,'' in \emph{Computer Vision and Pattern Recognition (CVPR)},
  2018, pp. 2345--2353.

\bibitem{kehl2017ssd6d}
W.~Kehl, F.~Manhardt, F.~Tombari, S.~Ilic, and N.~Navab, ``Ssd-6d: Making
  rgb-based 3d detection and 6d pose estimation great again,'' in
  \emph{Proceedings of the International Conference on Computer Vision (ICCV
  2017), Venice, Italy}, 2017, pp. 22--29.

\bibitem{chen20153dop}
X.~Chen, K.~Kundu, Y.~Zhu, A.~G. Berneshawi, H.~Ma, S.~Fidler, and R.~Urtasun,
  ``3d object proposals for accurate object class detection,'' in
  \emph{Advances in Neural Information Processing Systems}, 2015, pp. 424--432.

\bibitem{song2016deep}
S.~Song and J.~Xiao, ``Deep sliding shapes for amodal 3d object detection in
  rgb-d images,'' in \emph{The IEEE Conference on Computer Vision and Pattern
  Recognition (CVPR)}, June 2016.

\bibitem{liu2015higher}
J.~Liu, J.~Wang, T.~Fang, C.-L. Tai, and L.~Quan, ``Higher-order crf structural
  segmentation of 3d reconstructed surfaces,'' in \emph{IEEE International
  Conference on Computer Vision}, 2015.

\bibitem{yang2019std}
Z.~Yang, Y.~Sun, S.~Liu, X.~Shen, and J.~Jia, ``Std: Sparse-to-dense 3d object
  detector for point cloud,'' in \emph{Proceedings of the IEEE International
  Conference on Computer Vision}, 2019, pp. 1951--1960.

\bibitem{shi2019part}
S.~Shi, Z.~Wang, X.~Wang, and H.~Li, ``Part-a\^{} 2 net: 3d part-aware and
  aggregation neural network for object detection from point cloud,''
  \emph{arXiv preprint arXiv:1907.03670}, 2019.

\bibitem{liang2019multi}
M.~Liang, B.~Yang, Y.~Chen, R.~Hu, and R.~Urtasun, ``Multi-task multi-sensor
  fusion for 3d object detection,'' in \emph{Proceedings of the IEEE Conference
  on Computer Vision and Pattern Recognition}, 2019, pp. 7345--7353.

\bibitem{cvpr16chen}
X.~Chen, K.~Kundu, Z.~Zhang, H.~Ma, S.~Fidler, and R.~Urtasun, ``Monocular 3d
  object detection for autonomous driving,'' in \emph{IEEE CVPR}, 2016.

\bibitem{fu2018deep}
H.~Fu, M.~Gong, C.~Wang, K.~Batmanghelich, and D.~Tao, ``Deep ordinal
  regression network for monocular depth estimation,'' in \emph{Proceedings of
  the IEEE Conference on Computer Vision and Pattern Recognition}, 2018.

\bibitem{ren2019deep}
H.~Ren, M.~Elkhamy, and J.~Lee, ``Deep robust single image depth estimation
  neural network using scene understanding,'' pp. 37--45, 2019.

\bibitem{afifi2016object}
A.~J. Afifi and O.~Hellwich, ``Object depth estimation from a single image
  using fully convolutional neural network,'' in \emph{2016 International
  Conference on Digital Image Computing: Techniques and Applications
  (DICTA)}.\hskip 1em plus 0.5em minus 0.4em\relax IEEE, 2016, pp. 1--7.

\bibitem{zhang2015monocular}
Z.~Zhang, A.~G. Schwing, S.~Fidler, and R.~Urtasun, ``Monocular object instance
  segmentation and depth ordering with cnns,'' in \emph{Proceedings of the IEEE
  International Conference on Computer Vision}, 2015, pp. 2614--2622.

\bibitem{lee2019instance}
S.~Lee, S.~Im, S.~Lin, and I.~S. Kweon, ``Instance-wise depth and motion
  learning from monocular videos,'' \emph{arXiv preprint arXiv:1912.09351},
  2019.

\bibitem{han2015object}
J.~Han, D.~Zhang, G.~Cheng, L.~Guo, and J.~Ren, ``Object detection in optical
  remote sensing images based on weakly supervised learning and high-level
  feature learning,'' \emph{IEEE Transactions on Geoscience and Remote
  Sensing}, vol.~53, no.~6, pp. 3325--3337, 2015.

\bibitem{sangineto2019self}
E.~Sangineto, M.~Nabi, D.~Culibrk, and N.~Sebe, ``Self paced deep learning for
  weakly supervised object detection,'' \emph{IEEE transactions on pattern
  analysis and machine intelligence}, vol.~41, no.~3, pp. 712--725, 2019.

\bibitem{zhang2018adversarial}
X.~Zhang, Y.~Wei, J.~Feng, Y.~Yang, and T.~S. Huang, ``Adversarial
  complementary learning for weakly supervised object localization,'' in
  \emph{Proceedings of the IEEE Conference on Computer Vision and Pattern
  Recognition}, 2018, pp. 1325--1334.

\bibitem{zhang2018zigzag}
X.~Zhang, J.~Feng, H.~Xiong, and Q.~Tian, ``Zigzag learning for weakly
  supervised object detection,'' in \emph{Proceedings of the IEEE Conference on
  Computer Vision and Pattern Recognition}, 2018, pp. 4262--4270.

\bibitem{tang2018pcl}
P.~Tang, X.~Wang, S.~Bai, W.~Shen, X.~Bai, W.~Liu, and A.~L. Yuille, ``Pcl:
  Proposal cluster learning for weakly supervised object detection,''
  \emph{IEEE transactions on pattern analysis and machine intelligence}, 2018.

\bibitem{li2019stereo}
P.~Li, X.~Chen, and S.~Shen, ``Stereo r-cnn based 3d object detection for
  autonomous driving,'' in \emph{Proceedings of the IEEE Conference on Computer
  Vision and Pattern Recognition}, 2019, pp. 7644--7652.

\bibitem{qin2019tlnet}
Z.~Qin, J.~Wang, and Y.~Lu, ``Triangulation learning network: from monocular to
  stereo 3d object detection,'' in \emph{IEEE Conference on Computer Vision and
  Pattern Recognition (CVPR)}, 2019.

\bibitem{braun2016pose}
M.~Braun, Q.~Rao, Y.~Wang, and F.~Flohr, ``Pose-rcnn: Joint object detection
  and pose estimation using 3d object proposals,'' in \emph{2016 IEEE 19th
  International Conference on Intelligent Transportation Systems (ITSC)}.\hskip
  1em plus 0.5em minus 0.4em\relax IEEE, 2016, pp. 1546--1551.

\bibitem{pathak2017learning}
D.~Pathak, R.~Girshick, P.~Doll{\'a}r, T.~Darrell, and B.~Hariharan, ``Learning
  features by watching objects move,'' in \emph{Proceedings of the IEEE
  Conference on Computer Vision and Pattern Recognition}, 2017, pp. 2701--2710.

\bibitem{dai2016r}
J.~Dai, Y.~Li, K.~He, and J.~Sun, ``R-fcn: Object detection via region-based
  fully convolutional networks,'' in \emph{Advances in neural information
  processing systems}, 2016, pp. 379--387.

\bibitem{lim2011transfer}
J.~J. Lim, R.~R. Salakhutdinov, and A.~Torralba, ``Transfer learning by
  borrowing examples for multiclass object detection,'' in \emph{Advances in
  neural information processing systems}, 2011, pp. 118--126.

\bibitem{lampert2009learning}
C.~H. Lampert, H.~Nickisch, and S.~Harmeling, ``Learning to detect unseen
  object classes by between-class attribute transfer,'' in \emph{2009 IEEE
  Conference on Computer Vision and Pattern Recognition}.\hskip 1em plus 0.5em
  minus 0.4em\relax IEEE, 2009, pp. 951--958.

\bibitem{gustafsson2018automotive}
F.~Gustafsson and E.~Linder-Nor{\'e}n, ``Automotive 3d object detection without
  target domain annotations,'' 2018.

\bibitem{hoffman2014lsda}
J.~Hoffman, S.~Guadarrama, E.~S. Tzeng, R.~Hu, J.~Donahue, R.~Girshick,
  T.~Darrell, and K.~Saenko, ``Lsda: Large scale detection through
  adaptation,'' in \emph{Advances in Neural Information Processing Systems},
  2014, pp. 3536--3544.

\bibitem{spinello2012leveraging}
L.~Spinello and K.~O. Arras, ``Leveraging rgb-d data: Adaptive fusion and
  domain adaptation for object detection,'' in \emph{2012 IEEE International
  Conference on Robotics and Automation}.\hskip 1em plus 0.5em minus
  0.4em\relax IEEE, 2012, pp. 4469--4474.

\bibitem{sun2018pwc}
D.~Sun, X.~Yang, M.-Y. Liu, and J.~Kautz, ``{PWC-Net}: {CNNs} for optical flow
  using pyramid, warping, and cost volume,'' 2018.

\bibitem{mehta2018single}
D.~Mehta, O.~Sotnychenko, F.~Mueller, W.~Xu, S.~Sridhar, G.~Pons-Moll, and
  C.~Theobalt, ``Single-shot multi-person 3d pose estimation from monocular
  rgb,'' in \emph{2018 International Conference on 3D Vision (3DV)}.\hskip 1em
  plus 0.5em minus 0.4em\relax IEEE, 2018, pp. 120--130.

\bibitem{van1995python}
G.~Van~Rossum and F.~L. Drake~Jr, \emph{Python tutorial}.\hskip 1em plus 0.5em
  minus 0.4em\relax Centrum voor Wiskunde en Informatica Amsterdam, The
  Netherlands, 1995.

\bibitem{tensorflow2015-whitepaper}
M.~Abadi, A.~Agarwal, P.~Barham, E.~Brevdo, Z.~Chen, C.~Citro, G.~S. Corrado,
  A.~Davis, J.~Dean, M.~Devin, S.~Ghemawat, I.~Goodfellow, A.~Harp, G.~Irving,
  M.~Isard, Y.~Jia, R.~Jozefowicz, L.~Kaiser, M.~Kudlur, J.~Levenberg,
  D.~Man\'{e}, R.~Monga, S.~Moore, D.~Murray, C.~Olah, M.~Schuster, J.~Shlens,
  B.~Steiner, I.~Sutskever, K.~Talwar, P.~Tucker, V.~Vanhoucke, V.~Vasudevan,
  F.~Vi\'{e}gas, O.~Vinyals, P.~Warden, M.~Wattenberg, M.~Wicke, Y.~Yu, and
  X.~Zheng, ``{TensorFlow}: Large-scale machine learning on heterogeneous
  systems,'' 2015.

\bibitem{matthew2014vgg}
D.~Z. Matthew and F.~Rob, ``Visualizing and understanding convolutional
  networks,'' in \emph{European Conference on Computer Vision}, 2014, pp.
  818--833.

\bibitem{imagenet15}
O.~Russakovsky, J.~Deng, H.~Su, J.~Krause, S.~Satheesh, S.~Ma, Z.~Huang,
  A.~Karpathy, A.~Khosla, M.~Bernstein, A.~C. Berg, and L.~Fei-Fei, ``{ImageNet
  Large Scale Visual Recognition Challenge},'' \emph{International Journal of
  Computer Vision (IJCV)}, vol. 115, no.~3, pp. 211--252, 2015.

\bibitem{kin2015adam}
D.~P. Kingma and J.~Ba, ``Adam: A method for stochastic optimization,'' in
  \emph{International Conference for Learning Representations}, 2015.

\bibitem{Liu_2019_CVPR}
L.~Liu, J.~Lu, C.~Xu, Q.~Tian, and J.~Zhou, ``Deep fitting degree scoring
  network for monocular 3d object detection,'' in \emph{The IEEE Conference on
  Computer Vision and Pattern Recognition (CVPR)}, June 2019.

\bibitem{ma2019accurate}
X.~Ma, Z.~Wang, H.~Li, P.~Zhang, W.~Ouyang, and X.~Fan, ``Accurate monocular 3d
  object detection via color-embedded 3d reconstruction for autonomous
  driving,'' in \emph{Proceedings of the IEEE International Conference on
  Computer Vision}, 2019, pp. 6851--6860.

\bibitem{fu2018ordinal}
H.~Fu, M.~Gong, C.~Wang, K.~Batmanghelich, and D.~Tao, ``Deep ordinal
  regression network for monocular depth estimation,'' in \emph{Computer Vision
  and Pattern Recognition (CVPR)}, 2018.

\bibitem{manhardt2019roi}
F.~Manhardt, W.~Kehl, and A.~Gaidon, ``Roi-10d: Monocular lifting of 2d
  detection to 6d pose and metric shape,'' in \emph{Proceedings of the IEEE
  Conference on Computer Vision and Pattern Recognition}, 2019, pp. 2069--2078.

\bibitem{torrey2010transfer}
L.~Torrey and J.~Shavlik, ``Transfer learning,'' in \emph{Handbook of research
  on machine learning applications and trends: algorithms, methods, and
  techniques}.\hskip 1em plus 0.5em minus 0.4em\relax IGI Global, 2010, pp.
  242--264.

\bibitem{Chen20173D}
X.~Chen, K.~Kundu, Y.~Zhu, H.~Ma, S.~Fidler, and R.~Urtasun, ``3d object
  proposals using stereo imagery for accurate object class detection,''
  \emph{IEEE Transactions on Pattern Analysis and Machine Intelligence}, 2017.

\bibitem{guizilini20203d}
V.~Guizilini, R.~Ambrus, S.~Pillai, A.~Raventos, and A.~Gaidon, ``3d packing
  for self-supervised monocular depth estimation,'' in \emph{Proceedings of the
  IEEE/CVF Conference on Computer Vision and Pattern Recognition}, 2020, pp.
  2485--2494.

\bibitem{godard2019digging}
C.~Godard, O.~Mac~Aodha, M.~Firman, and G.~J. Brostow, ``Digging into
  self-supervised monocular depth estimation,'' in \emph{Proceedings of the
  IEEE international conference on computer vision}, 2019, pp. 3828--3838.

\end{thebibliography}

\begin{IEEEbiography}[{\includegraphics[width=1in,height=1.25in,clip,keepaspectratio]{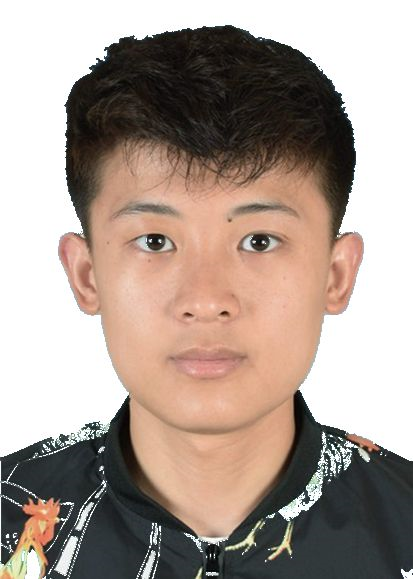}}]{Zengyi Qin} is a first year graduate student in the Department of Aeronautics and Astronautics at MIT, advised by Prof. Chuchu Fan. He received the B.E. degree (with honor) in Electronic Engineering at Tsinghua University, China. During his undergraduate study, he also spent 1 year at Microsoft Research Asia and did research internship at Stanford University. His research interests include safe autonomous systems, computer vision and robotics.
\end{IEEEbiography}

\begin{IEEEbiography}[{\includegraphics[width=1in,height=1.25in,clip,keepaspectratio]{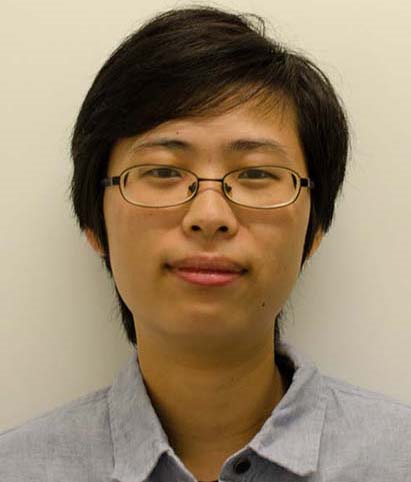}}]{Jinglu Wang} received her bachelor degree in Computer Science and Technology from Fudan University, Shanghai, China, in 2011, and PhD degree in Computer Science and Engineering the Hong Kong University of Science and technology, Hong Kong, China, in 2016. She is currently a senior researcher of Media Computing Group, Microsoft Research Asia. Her research interests include 3D reconstruction, 3D object detection, and video segmentation.
\end{IEEEbiography}

\begin{IEEEbiography}[{\includegraphics[width=1in,height=1.25in,clip,keepaspectratio]{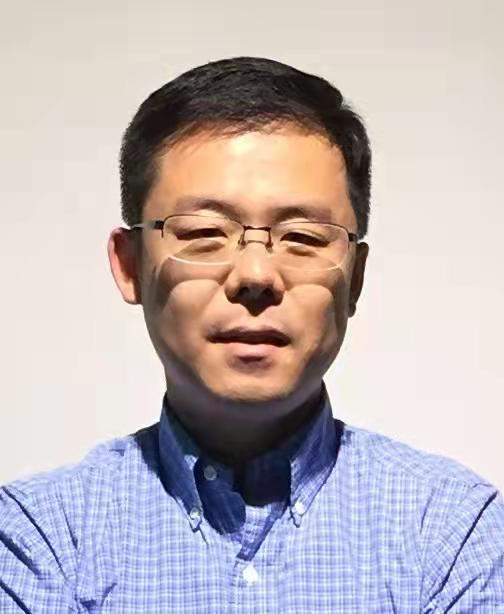}}]{Yan Lu} received his Ph.D. degree in computer science from Harbin Institute of Technology, China. He joined in Microsoft Research Asia in 2004, where he is now the Partner Research Manager of Media Computing Group, leading the development of core technologies in the fields of real-time communication, computer vision, video analytics, audio enhancement, virtualization, and mobile-cloud computing. From 2001 to 2004, Yan Lu was the team lead of video coding group in the JDL Lab, Institute of Computing Technology, China. From 1999 to 2000, he was with the City University of Hong Kong as a research assistant. His research interests include image and video coding, computer vision, audio and speech, multimedia system, networking, and remote computing. Yan Lu has published over 100 papers and holds over 30 granted US patents in the field of multimedia and computer vision.
\end{IEEEbiography}

\end{document}